\documentclass[preprint,12pt]{elsarticle}

\usepackage{amssymb}
\usepackage{amsmath}

\journal{Journal of Manufacturing Systems}

\usepackage{lineno}
\usepackage{hyperref}
\usepackage[capitalize,nameinlink,noabbrev]{cleveref}
\usepackage[mode=buildnew]{standalone} 
\usepackage{booktabs}
\usepackage{multirow}
\usepackage{adjustbox}

\usepackage[acronym]{glossaries}
\glsdisablehyper

\newacronym{ai}{AI}{Artificial Intelligence}
\newacronym{am}{AM}{Additive Manufacturing}
\newacronym{it}{IT}{Information Technology}
\newacronym{gnn}{GNN}{Graph Neural Network}
\newacronym{cnn}{CNN}{Convolutional Neural Network}
\newacronym{dl}{DL}{Deep Learning}
\newacronym{doi}{DOI}{Digital Object Identifier}
\newacronym{ml}{ML}{Machine Learning}
\newacronym{ot}{OT}{Operational Technology}
\newacronym{slt}{SLT}{System Level Test}
\newacronym{ph}{PH}{Persistent Homology}
\newacronym{pi}{PI}{Persistence Image}
\newacronym{som}{SOM}{Self Organizing Map}
\newacronym{tda}{TDA}{Topological Data Analysis}
\newacronym{umap}{UMAP}{Uniform Manifold Approximation and Projection}

\usepackage{xspace}
\newcommand{\tda}{\gls{tda}\xspace}
\newcommand{\ml}{\gls{ml}\xspace}
\newcommand{\ph}{\gls{ph}\xspace}
\newcommand{\umap}{\gls{umap}\xspace}
\newcommand{\indfour}{Industry 4.0\xspace}
\usepackage{color}
\usepackage{mathtools}

\newcommand{\ignore}[1]{}
\newcommand{\uram}[1]{\marginpar{\tiny\textcolor{blue}{UraM: #1}}}

\renewcommand{\uram}{}

\begin{document}

\begin{frontmatter}



\title{Topological Data Analysis in Smart Manufacturing: State of the Art and Future Directions}

\author[fhs,plus]{Martin Uray}
\ead{martin.uray@fh-salzburg.ac.at}
\author[albany]{Barbara Giunti}
\author[tug]{Michael Kerber}
\author[fhs]{Stefan Huber}
\affiliation[fhs]{organization={Josef Ressel Centre for Intelligent
    and Secure Industrial Automation, Salzburg University of Applied Sciences},
            addressline={Urstein Sued 1},
            postcode={5412},
            city={Puch/Hallein},
            country={Austria}}
\affiliation[plus]{organization={Department of Artificial Intelligence and
Human Interfaces, Paris-Lodron-University of Salzburg},
            addressline={Jakob-Haringer-Str. 2},
            postcode={5020},
            city={Salzburg},
            country={Austria}}
\affiliation[albany]{organization={Department of Mathematics and Statistics,
University at Albany},
            addressline={1400 Washington Avenue},
            city={Albany},
            postcode={12222},
            state={NY},
            country={USA}}
\affiliation[tug]{organization={Institute of Geometry, Graz University of Technology},
            addressline={Kopernikusgasse 24},
            postcode={8010},
            city={Graz},
            country={Austria}}

\begin{abstract}
\tda is a discipline that applies algebraic topology techniques
to analyze complex, multi-dimensional data.
Although it is a relatively new field, \tda has been widely and successfully
applied across various domains, such as medicine, materials science, and
biology.
This survey provides an overview of the state of the art of \tda within a
dynamic and promising application area: industrial manufacturing and
production, particularly within the \indfour context.
%
We have conducted a rigorous and reproducible literature search focusing on \tda
applications in industrial production and manufacturing settings.
The identified works are categorized based on their application
areas within the manufacturing process and the types of input data.
We highlight the principal advantages of \tda tools in this
context, address the challenges encountered and the future potential of
the field.
Furthermore, we identify \tda methods that are currently underexploited in
specific industrial areas and discuss how their application could be
beneficial, with the aim of stimulating further research in this field.
%
This work seeks to bridge the theoretical advancements in \tda with the
practical needs of industrial production.
Our goal is to serve as a guide for practitioners and researchers applying \tda
in industrial production and manufacturing systems.
We advocate for the untapped potential of \tda in this domain and encourage
continued exploration and research.

\end{abstract}



\end{frontmatter}



    \section{Introduction}\label{sec:introduction}
\indfour
represents the fourth industrial revolution, marked by the integration of
digital and physical technologies.
It is reshaping manufacturing and facilitating the
creation of smart production systems.
Such systems are distinguished by their geographic distribution,
interconnectedness, and capability for autonomous decision-making to swiftly
adapt to changes in production requirements~\cite{bai2020}.
Traditionally, these adaptations were managed by operators, a process requiring
extensive training and multiple attempts to produce acceptable goods.

Topology, a branch of mathematics, focuses on studying the
properties of objects that remain unchanged under continuous deformations.
\glsfirst{tda}, an emerging field at the confluence of data analysis, computer
science, and algebraic topology, leverages the underlying topological and geometrical structure
of datasets to uncover features not detectable with other analysis methods.
This approach enables the analysis and visualization of topological
structures such as connected components, loops, and cavities, and has been proved
efficient across various applications including anomaly
detection~\cite{stolz2020}, image processing~\cite{cavinato2022}, and genome
sequencing~\cite{rabadan2019a}, as well as in sectors like the chemical
industry~\cite{smith2021}, finance~\cite{ruiz-ortiz2023},
aviation~\cite{li2019}, and physics~\cite{hamilton2022}.

The shared challenges and characteristics across different fields highlight
the substantial promise of \tda in industrial production contexts.
Yet, the integration of \tda within industrial manufacturing is notably limited.
We claim that \tda is particularly effective for intelligent production
systems, as it facilitates the extraction of valuable insights from complex data
generated by sensors and various devices, thus supporting enhanced
decision-making.
However, its comprehensive potential in the sphere of \indfour has not been
fully realized.

\subsection{Contribution \& Outline}\label{subsec:contribution-outline}

This paper offers a comprehensive review of the current literature on the
applications of \gls{tda} in industrial production and manufacturing.
It aims to bridge the existing gap in mutual awareness between theorists in
\gls{tda} and practitioners in industrial production, facilitating the exchange of ideas and methods between these
domains. 
Since the theoretical nature of \tda may be off-putting, we provide a hands-on description of its tools 
and exemplify them with industrial applications.

This paper makes the following four contributions:
\begin{enumerate}
    \item It provides an overview of the current literature on \gls{tda}
          applications within industrial production and manufacturing;
    \item It demonstrates how \gls{tda} methods are currently being integrated
          into smart manufacturing;
    \item It highlights areas within smart manufacturing where \gls{tda} methods
          are underexplored;
    \item It outlines some guidelines for off-the-shelf uses of \gls{tda} in industrial production and manufacturing.
\end{enumerate}

The structure of this paper is as follows: \Cref{sec:industry-4.0} introduces
the necessary background and key terminology related to \indfour.
\Cref{sec:tda} presents \gls{tda} and its principal tools.
The methodology of the survey is outlined in \Cref{sec:survey}, followed by
the presentation of results in \Cref{sec:results}, which includes a
comprehensive discussion for each identified application domain.
\Cref{sec:discussion} interprets the findings and suggests directions for
future research.
The paper concludes in \Cref{sec:conclusion}.

\subsection{Related Work}\label{subsec:related-work}

The application scope of \gls{tda} has expanded so significantly,
encompassing over 450 papers to date, that it has become virtually impossible
for a single researcher to keep up with all the advancements.
This proliferation raises the question of how to structure the interface
between \gls{tda} and its application domains to make it accessible to both
theorists and practitioners.

The platform DONUT, introduced by Giunti et al.~\cite{giunti2022},
offers a search engine designed to facilitate the exploration of \tda
applications.
Additionally, scientific surveys play a crucial role in organizing
this knowledge, summarizing and contrasting different approaches applying
\gls{tda} in practical settings.
There exists a considerable collection of such surveys
(e.g.~\cite{ghrist2007,edelsbrunner2008,kerber2016,edelsbrunner2017,munch2017,perea2018,vejdemo-johansson2013})
and textbooks (e.g.~\cite{edelsbrunner2010,carlsson2021a,oudot2015}) that generally
focus on explaining the theory and showcasing exemplary applications.
Nevertheless, we found none focusing on industrial manufacturing.
There are, however, reviews on \gls{tda} applications in other fields, such as
Topological Time Series~\cite{perea2019} or Topological
\gls{ml}~\cite{hensel2021,pun2022,papamarkou2024}, as well as reviews on applications in 
manufacturing without a \gls{tda} focus.

Capodieci et al.~\cite{capodieci2017} review \gls{ml} and other data analysis
methods for yield optimization in semiconductor manufacturing, briefly
mentioning Mapper's application.
Gao et al.~\cite{gao2020a} discuss the potential of \gls{tda} methods, including
Mapper, for Big Data analytics in smart factories.
Kounta et al.~\cite{kounta2022} and Liewald et al.~\cite{liewald2022a} highlight
\gls{tda}'s potential in machining chatter detection and metal forming,
respectively.

Additionally, Big Data's relevance to \indfour is underscored in a survey by
Snasel et al.~\cite{snasel2017a}, aiming to bridge theoretical and engineering
disciplines.
Wang~\cite{wang2020d} discusses \gls{ph} and Mapper in 3D
printing, a key technology in additive manufacturing.


    \section[Smart Manufacturing in I4.0]{Smart Manufacturing in Industry~4.0}
\label{sec:industry-4.0}

\subsection[Industry 4.0]{Industry~4.0}
\label{subsec:industry4.0}

The term \indfour was coined in the early 2010's by the German government in
order to foster and drive the so-called ``fourth industrial revolution''
(see~\cite{schwab2016} for details).
This movement is driven by the need for higher flexibility in production,
the operation of machines that are more adaptive to the requirements of the production,
and
a smarter, more autonomous way of operating machines, production lines, factories and even whole supply chains.
This enables production paradigms like lot-size-one production and mass customization.
However, it also fosters the optimization of
production scenarios not only per machine but on the entire value chain, eventually
enabling new business and operational models as well.
This vision gives rise to various other terms along with \indfour, like ``smart
factory'' or ``cognitive factory''.

However, \indfour is not only about manufacturing of products only, but its context is much wider.
Indeed,\ \cite{schwab2016} argue that this paradigm enables more innovative and
revolutionary products per se by incorporating other technologies that are
occurring simultaneously.
These technologies include gene sequencing, nanotechnologies, renewables, quantum computing,
and much more~\cite{schwab2016}.

Hermann et al.~\cite{hermann2016} identified the following four design
principles of \indfour systems:
\begin{itemize}
    \item \emph{Interconnection}: all components, like sensors, machines, and
    even humans, are connected with each other.

    \item \emph{Information Transparency}: the information about all
    components are transparent within the system.
    This enables operators to make intelligent and well-informed decisions.

    \item \emph{Technical Assistance}: the technological facilities assist
    humans in decision-making, support in problem-solving, and help or take
    over hard or unsafe tasks.

    \item \emph{Decentralized Decisions}: decisions are not made by a central
    instance, rather they are made ``on the edge''.
    Cyber-physical systems are able to make decisions on their own, based on
    the information they have.
    Exceptions to this rule are in inferences or conflicting aims, delegated
    to a higher instance.
\end{itemize}

A review by Erboz~\cite{erboz2017} identified main components of systems of
\indfour: Big Data and Analytics,
Autonomous Robots, Simulation, Horizontal and Vertical
System Integration, Industrial Internet of Things (IIoT),
Cloud, \emph{Cybersecurity}, \emph{Additive Manufacturing}, and
Augmented Reality (AR).
When talking about \emph{Cybersecurity} in the context of \indfour, or \gls{ot}
in general, we further refer to the term \gls{ot} security~\cite{stouffer2022}.
The term \emph{Additive Manufacturing} refers to the technology of \emph{3D Printing} in an industrial
production context.
Here, three-dimensional objects are created layer-wise by
deposing material in a computer-controlled process~\cite{ngo2018}.

\subsection{Production and manufacturing}\label{subsec:production-and-manufacturing}

In the literature, the terms \emph{Manufacturing} and \emph{Production} are
used for the manufacturing process of producing goods.
Depending on the domain also other terms are found, e.g.,
\emph{Fabrication} in semiconductors~\cite{kalpakjian2014}.
Despite slight semantic differences, we use the term
\emph{Production} and \emph{Manufacturing} as umbrella terms for these three
terms within this survey.

\begin{figure}
    \centering
    \includegraphics[width=0.5\columnwidth]{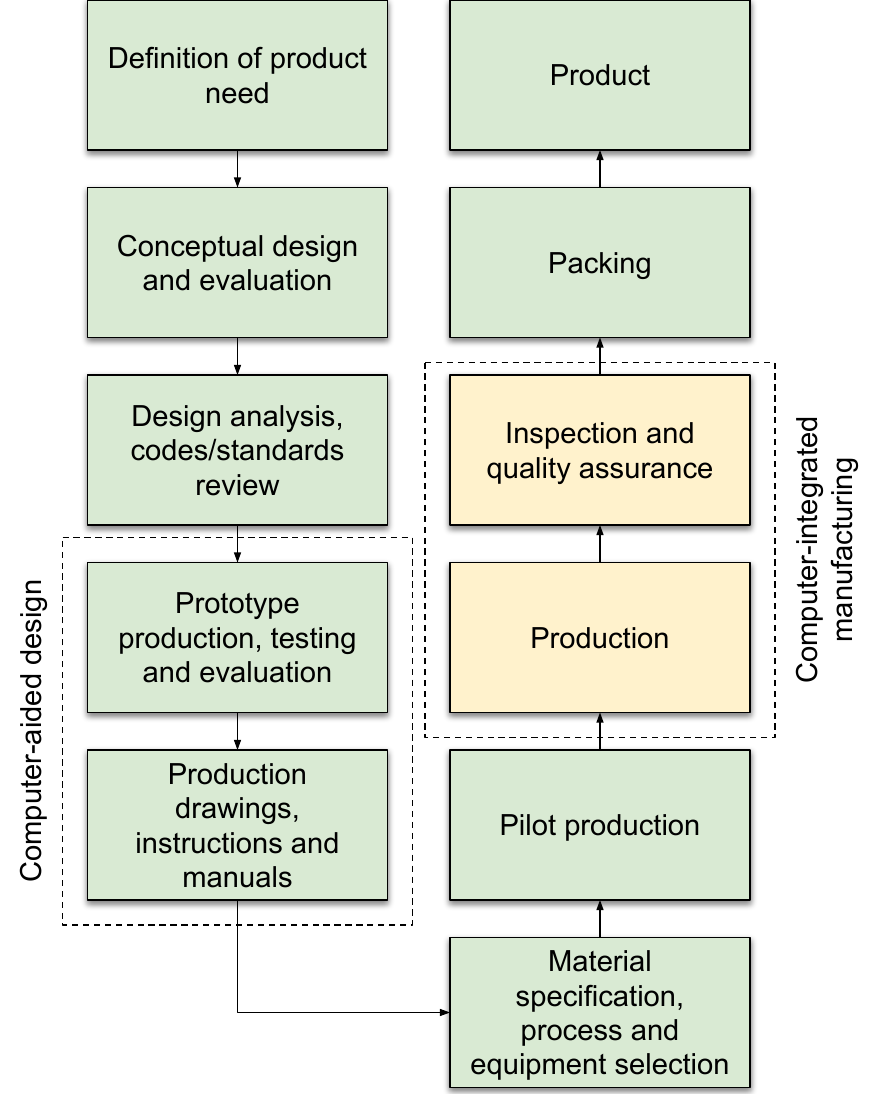}
    \caption{The manufacturing engineering process with its stages along
    product production, beginning from the productions definition to its
    final, mass-produced artifact. Feedback connections omitted for the
    sake of readability.
    Illustration adapted from~\cite{kalpakjian2014}.}
    \label{fig:production_process}
\end{figure}

The manufacturing of a product involves a sequence of process steps, applied by
industrial machines in a production line.
\Cref{fig:production_process} illustrates the stages of the manufacturing
engineering process sequentially.
At the beginning of a design process, the definition of the product
requirements have to be identified, followed by a conceptional design and
evaluation of the said.
Based on that, a prototype is created, enabling the
creation of schematics for industrial reproduction, where ``industrial''
typically means in a repeatable, efficient, and effective way.
These schematics,
in combination with the requirements of the product, define the specification
for the selection of materials, processes, and production equipment.
The
production itself is then accompanied and finished by an inspection and quality
assurance before the products are packed.

Manufacturing- or Production Engineering describes the branch of engineering
working on the entire process of manufacturing.
Among others, the planning and optimization of the production process are
subject of interest to this discipline~\cite{matisoff1986}.

In general, smart manufacturing in \indfour poses new challenges to the
domain, compared to traditional manufacturing.
Here, additional strategies and technologies are used to improve the
manufacturing process, to fulfill the needs for the integration into \indfour.
An overview on the technologies and architectures used for smart
manufacturing systems is given in~\cite{chen2018a}.

    \section{Topological Data Analysis}\label{sec:tda}

We consider three main methods in \gls{tda}:
Mapper~\cite{singh2007}, \gls{ph}~\cite{edelsbrunner2010,huber2021}, and \gls{umap}~\cite{mcinnes2020}.
\gls{ph} is the prevalent method in \gls{tda}, Mapper a topologically guided
clustering algorithm, and \gls{umap} a topology-driven dimensionality reduction
method.
Common to all of them is that (i) the data at hand is converted into a
suitable (geo)metric representation and (ii) topological properties are
analyzed to extract (novel) information from the given representation.
In a remote analogy, where the Fourier analysis tells us something about
spectral properties of a signal, \gls{tda} extracts information about the shape
of various kind of data (e.g., loops, tunnels, cavities, components, et cetera).

\begin{figure*}[h!]
    \centering
    \includegraphics[width=0.8\textwidth]{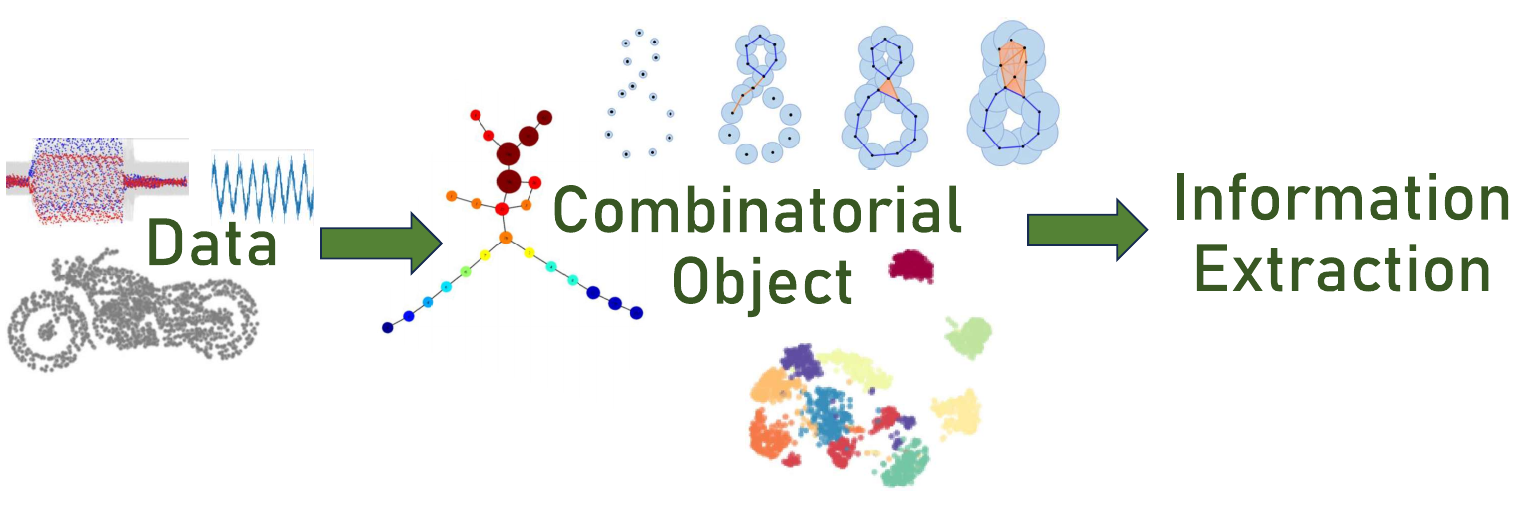}
    \caption{General \gls{tda} pipeline. Images of the time series (real and simulated) are taken from \cite[Figures 4 and 9]{yesilli2022a}, respectively. The simplicial complexes and the point cloud images come from \cite[Figures 1 and 3]{wong2021a}, respectively. The graph and the clustered graphs are taken from \cite[Figure 5]{singh2007} and \cite[Figure 2]{mcinnes2020}, respectively.}
    \label{fig:tda_pipeline}
\end{figure*}

The high-level pipeline of the three methods is depicted in
\Cref{fig:tda_pipeline}: input data is transformed into a combinatorial
object, which is then used to extract and interpret information.
This is done either directly (Mapper and \gls{ph}) or by feeding
its (vectorized) output to a \gls{ml} pipeline (all three methods).

During the stage of data preprocessing, data usually is manipulated by a set of
parameters.
The three aforementioned methods deal with parameters -- i.e., the step from
input data to a combinatorial object -- in different ways.
First, Mapper assembles parameters (and their values) in
different groups and then clusters the inputs accordingly.
This grouping can find latent relations in the dataset.
Similarly, \gls{umap} is a dimensionality reduction method.
By projecting the data into lower-dimensional ambient space, redundant or not relevant information in the data is
compressed.
This enables the input to be analyzed more easily.
On the other hand, \gls{ph} overcomes the need of choosing parameter values by
considering all possible values at the same time. 
Instead of outputting information about the shapes in the data for a single parameter value, \gls{ph} keeps track of how the shapes evolve depending on the variation of parameter(s).
\gls{ph} is especially suitable for automated production since it removes the need to cherry-pick a threshold (see for
example~\cite{casolo2022}).

A considerable attribute common to all \gls{tda} methods is that they are
applicable to virtually any scenario: as long as the data can be converted to
some general formats, the methods are agnostic to the data's source.
Therefore, \gls{tda} can be applied to any domain that has data in a suitable input form or that can be transformed into it, such as point clouds, meshes, polygonal shapes, time series, graphs, pictures, and more.
Domain-specific knowledge is only needed at a later stage to interpret the
results.


\begin{figure}[h!]
    \centering
    \includegraphics[width=0.5\columnwidth]{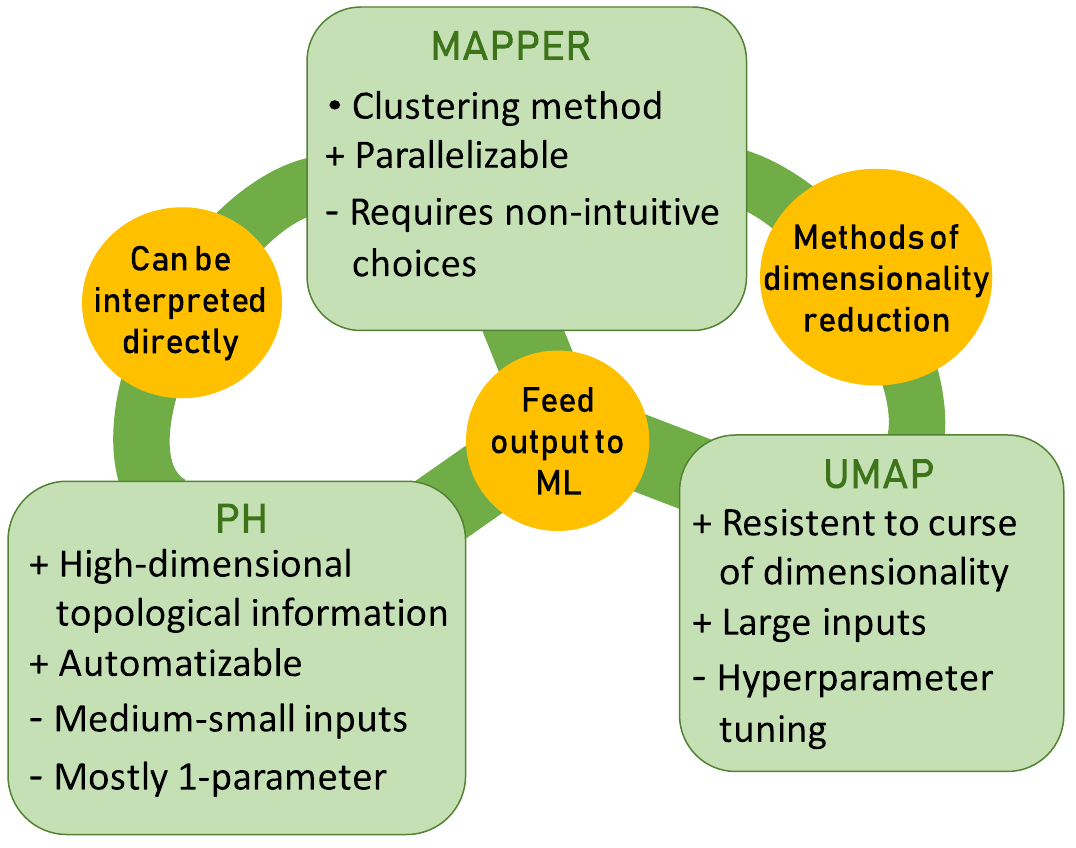}
    \caption{Scheme of the described \gls{tda} methods, with pro, contra, and (dis)similarities.}
    \label{fig:tda_pro_con}
\end{figure}

In \Cref{fig:tda_pro_con}, we summarize the advantages and disadvantages of the
three aforementioned methods and highlight the dissimilarities and shared characteristics.

\subsection{Mapper}\label{subsec:mapper}

Mapper~\cite{singh2007} reduces the data dimension by clustering
the input points in a graph, the so-called Mapper Graph.
The graph's vertices represent the clusters and the edges signify that two clusters share some elements.
This representation can further be analyzed for its topological or geometrical
properties, such as the connected components, loops, and flares. 

\begin{figure}[h!]
    \centering
    \includegraphics[width=0.75\columnwidth]{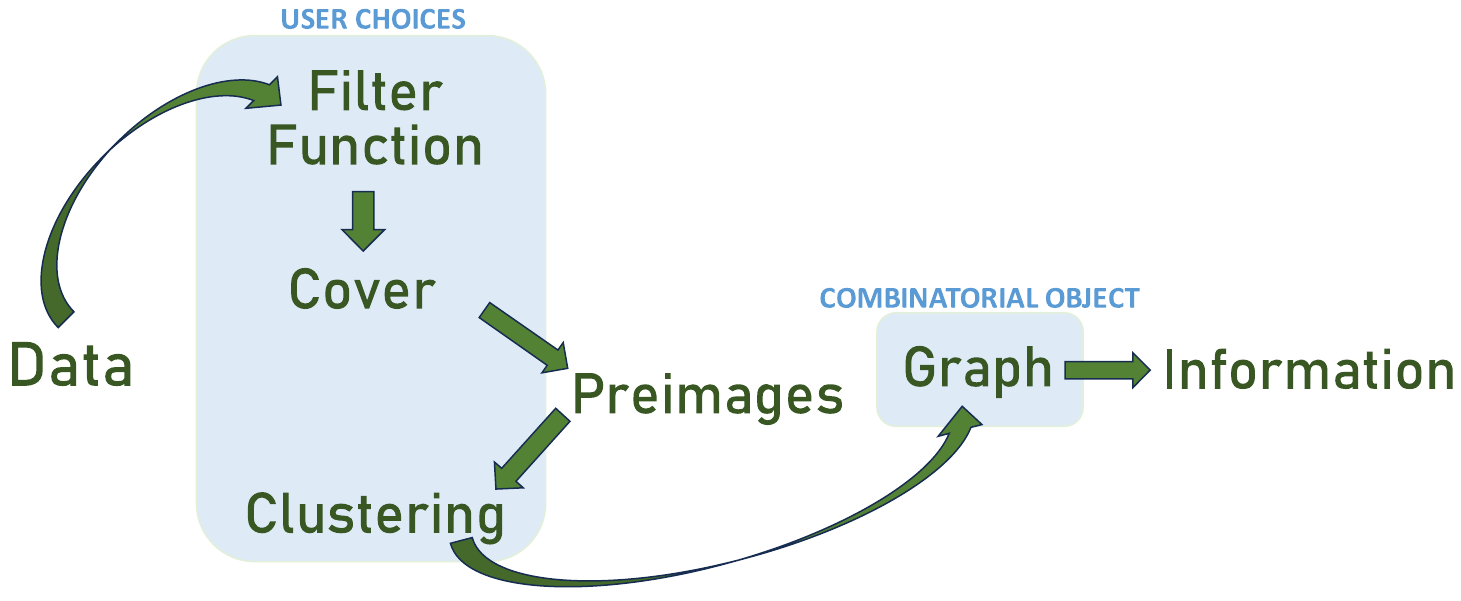}
    \caption{Mapper illustrated differently, indicating the
    choices the user has to make.}
    \label{fig:mapper_pipeline}
\end{figure}

\Cref{fig:mapper_pipeline} outlines a high-level overview of the Mapper
pipeline, highlighting the necessary choices to be made.
In detail, three choices are required:
\begin{enumerate}
    \item A real-value function $f\colon X \to \mathbb{R}$ over the input data,
        called \emph{filter} or \emph{lens function},
    \item An interval cover, \emph{cover} for brevity, of the image of $f$ (i.e.,
        a collection of possibly overlapping intervals whose union covers the
        image of $f$),
    \item A clustering algorithm.
\end{enumerate}

These choices are highly application-dependent and different choices usually
result in different outputs (and therefore in different interpretations).

The filter function may reflect geometrical properties of the data (for
example, a density estimator) or some wanted feature the user is interested
in (for example electrical properties).
In theory, Mapper works for general filter functions, e.g.,
maps to the circle $S^1$ or $\mathbb{R}^d$.
In practical applications, the filter function is typically a real-valued
function due to its simplicity and proven efficiency.

\begin{figure}[h!]
    \centering
    \includegraphics[width=0.75\textwidth]{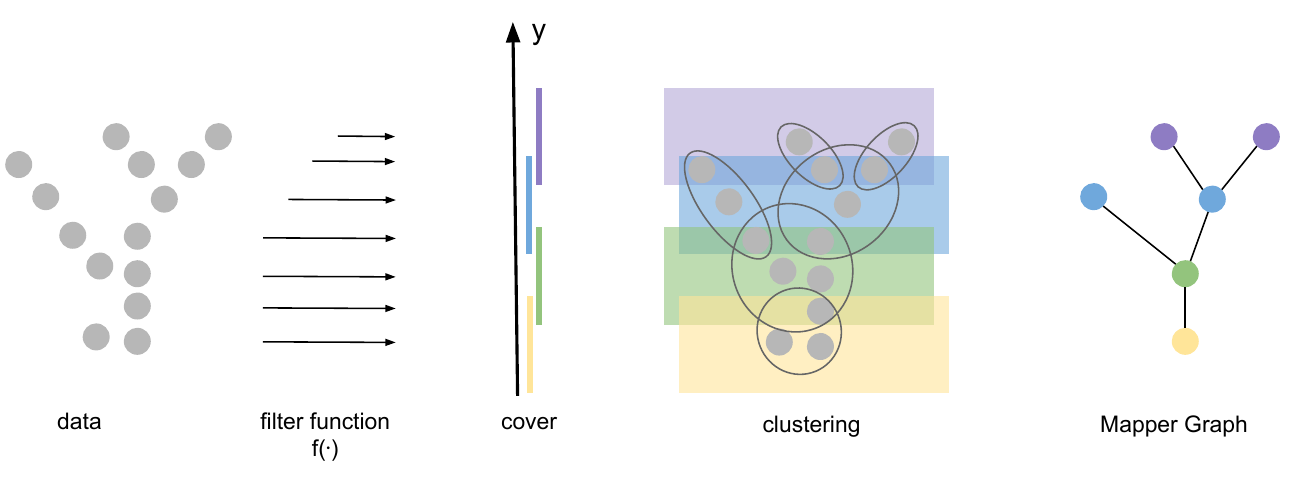}
    \caption{Graphical illustration of the Mapper Algorithm.}
    \label{fig:mapper_algorithm}
\end{figure}

We now describe the steps of Mapper (illustrated in
\Cref{fig:mapper_algorithm}):
first, the input data $X$ is reduced in dimension using the filter function $f$.
This function projects the input points into a low-dimensional space.
Then one chooses a cover $\mathcal{I}$ of the image of $f$ in the low-dimensional space. 
The idea is that the projection can facilitate the grouping of points by removing redundant information. 
This grouping, i.e.\ the cover, is then lifted back to the input points: one takes the preimage of each interval in the cover, i.e.,
the collection of all data $x \in X$ such that $f(x)\in \mathcal{I}$ for an interval $\mathcal{I}$ in the cover.
In other terms, all data points that are mapped to the same interval in the cover are grouped together in the domain.
Note that, since the intervals in the cover may overlap (and they usually do), a point $x \in X$ can belong to the preimages of several intervals.
Each of these preimages is clustered according to the chosen clustering algorithm.
Finally, each of these clusters is represented as a vertex, and two vertices have an edge between them if there is at least one $x\in X$ that belongs to both corresponding clusters.

The Mapper graph is used for explorative data analysis: for example, one looks
for \emph{flares} in the graph, that is, subpopulations of vertices that are
connected to each other across several intervals and disconnected from all others vertices.
One then analyzes these subpopulations (potentially with traditional data
analysis methods) to find a reason for their distinctiveness. 
For some practical examples outside industry, see~\cite{lum2013}.
Moreover, the Mapper graph can be feed to a \gls{ml} pipeline.
Implementations of Mapper are freely available (see, for
example,\ \cite{veen2019,tauzin2021}).

To give an idea of how the method works in practice, we now describe a practical
example taken from industrial manufacturing (c.f. \Cref{sec:results}).
The authors of~\cite{rivera-castro2019d} develop a demand-forecasting system of
electronic components that uses Mapper to identify which forecasting model
better predict the demand of a given component.
This system is constructed in two steps: first a Mapper Graph is built using
a training subset of components, and then the most fitting model for a new
component is selected using this graph.

The input data are time series encoding the demand of a given electronic component over a certain time span.
A small fraction of the data is used as the training set.
This fraction is preprocessed by computing the features describing each time series and its best forecasting model.
The collection of time series features is the actual input for Mapper, which
maps them using the first principal component as a filter function.
After covering the image and clustering the preimages, we obtain a graph
whose clusters represent electronic components whose demands behave similarly.
These clusters are labelled with the forecasting models that works best for
(the majority of) the electronic components in it.
The resulting Mapper graph can then be used to predict the best model
for new components.

This prediction of a new, unknown component is performed as follows:
the components features are calculated (as for the training sets). 
Then a $k$-nearest neighbor search is performed on the Mapper graph to find the nodes with similar features.
The cluster with the lowest distance to the new component is assigned to it.
The forecasting model of the assigned cluster is then applied to the new
component.

In practice, the major obstacles to the use of Mapper are the choices of the filter function $f$, cover, and clustering algorithm:
the interpretability of the outcome entirely depends on these choices. 
While a few standard choices are known, a domain expert is usually required to get meaningful insights out of the Mapper pipeline.
Indeed, the steps of the Mapper algorithm have very few mathematical constrains (the filter function can be anything, and the cover 
needs only to account for all points and consist of connected ``pieces'').
This in turn means that there are many degrees of freedom in applying this method, and that the correct choices are simply 
the ones that give better or more complete answers. 

Nevertheless, Mapper is a powerful, versatile method that can reveal covert connectivity in datasets.

\subsection{Uniform Manifold Approximation and Projection (UMAP)}\label{subsec:umap}

\begin{figure}[h!]
    \centering
    \includegraphics[width=0.75\columnwidth]{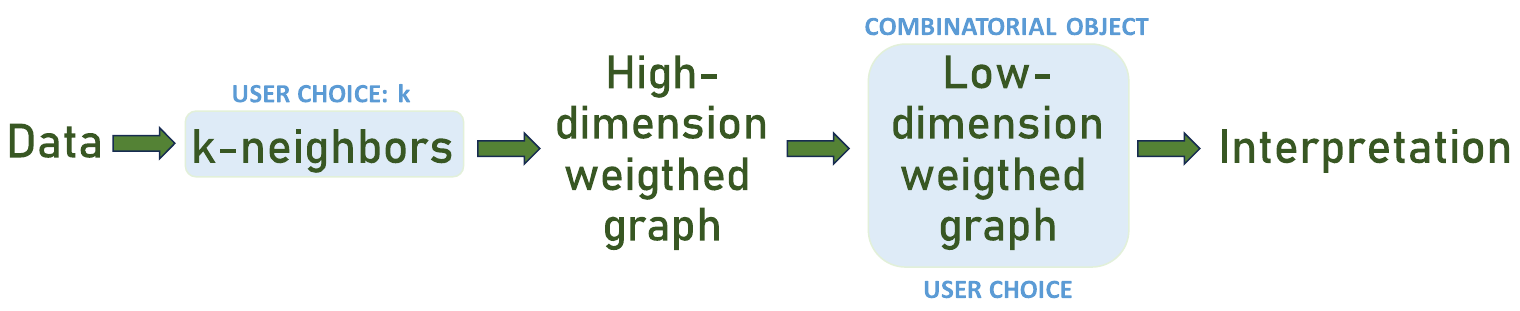}
    \caption{\gls{umap} pipeline. High/low-dimension refers to the ambient
                space where the graph lives.}
    \label{fig:umap_pipeline}
\end{figure}

\gls{umap}~\cite{mcinnes2020}, as Mapper, is a topologically driven dimensionality reduction method.
It works by building two (weighted) graphs.
The first graph is embedded in the same ambient dimension as the data and
approximates the geodesic distances of the underlying manifold.
The second graph approximates the first graph in a much lower ambient dimension.
The data's embedding in this lower dimension is then used for further analysis.
\Cref{fig:umap_pipeline} illustrates \gls{umap} schematically.

The first graph is constructed in several steps:
First, the $k$-nearest neighbors of each point $x_i$ in the input data $X$
are computed.
Here, $k$ is a parameter chosen by the user (different values of $k$ may result in different outcomes).
Then the geodesic distance from $x_i$ to each of its neighbor is approximated.
All distances for each input point $x_i$ are aggregated.
These distances provide an incomplete and conflicting collection.
Indeed, take two points $x_i, x_j \in X$.
The approximated distance from $x_i$ to $x_j$, computed while processing $x_i$,
is not necessarily the same as the one from $x_j$ to $x_i$, computed while processing $x_j$.
Moreover, while studying the $k$-nearest neighbors of $x_i$, we get the
approximated distance from it to $x_j$ and $x_h$, yet we have no information
about the distance between $x_j$ and $x_h$.
To overcome this issue, the authors use \emph{fuzzy structures} and directed graphs.
Fuzzy structures are a way to weight the membership to a set (instead of either being in or out, an element can belong to a set \emph{fuzzily}).
The membership strength is given, in this case, by the approximated distances. 
The directionality in the graph takes care of the asymmetry of the
distances.
We omit further technical details, which are not relevant for this survey (see~\cite{mcinnes2020} for details).

Since the distances are approximated from each point independently, the computation can be parallelized and it scales well with large inputs.
Another advantage of this construction is its robustness against the
curse of dimensionality, given by the fact that the distances are approximate locally.

The second graph is constructed by projecting the first graph into a lower dimensional manifold.
This projection is achieved via a \emph{force directed graph layout} algorithm, which
requires the user to choose two further hyperparameters.
Contrary to what happen with other methods of dimensionality reduction, such
as t-distributed stochastic neighbor embedding (tSNE)~\cite{maaten2008}, an
arbitrary target dimensionality can be selected.
The resulting second graph encodes the original local distances from the
input, but lives in a lower dimensional ambient space.
This final representation can be more easily
processed with other data analysis methods, in particular \gls{ml}.

There are several free implementations available of \gls{umap}, such
as~\cite{mcinnes2018,narayan2021,sainburg2021}.

The work by Hsu et al.~\cite{hsu2022a} exemplifies the application of \gls{umap}.
They analyze failures in \glspl{slt} on integrated circuits.
The need for robust interpretative methodologies is driven by the increasing
transistor density and complexity in circuit designs, where the high-dimensional nature of the associated data poses a significant
challenge.
To address this, Hsu et al.~\cite{hsu2022a} project high-dimensional data into a lower-dimensional space using
\gls{umap}.
The obtained lower-dimensional representation is analyzed with a
clustering algorithm, isolating clusters of systematically failing \glspl{slt}.
These clusters are then leveraged to construct a decision tree, enabling precise
identification of failures in individual components. 
The clustering algorithm is very efficient but sensitive to the curse of dimensionality.
Therefore, applying it directly to the data would lower the performances.

Contrary to the outputs of Mapper and \gls{ph}, the output of
\gls{umap} is usually not immediately interpretable, but can be used for further analysis, e.g., using \gls{ml}.
In conclusion, especially since it scales well with bigger input, and its robustness against the curse of dimensionality, \gls{umap} is a powerful tool for dimensionality reduction that has proven to be very successful in practice, outside and inside industry.

\subsection{Persistent Homology (PH)}\label{subsec:ph}

Many data preprocessing techniques require a fixed choice for the value of one or more parameters.
The key idea of \gls{ph} is to use topological methods to study how features
in data evolve for different choices of the parameter.
For instance, rather than choosing a fixed threshold for binary thresholding
in image analysis, we consider the sequence of results over a sweep of
thresholds.
Or instead of considering clusters of a point cloud by investigating the
union of balls of fixed radii centered at the points, we consider the
evolution of this union over a sweep of radii.
This not only allows us to avoid the choice of a parameter, but also to
extract more information from the data, by adding a dimension to the analysis.

\begin{figure}[h!]
    \centering
    \includegraphics[width=0.75\columnwidth]{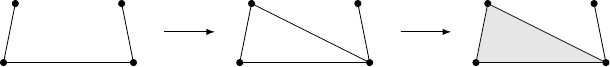}
    \caption{Three simplicial complexes, each including into the following forming a filtration.}
    \label{fig:simplicial_complex}
\end{figure}

In the case of \gls{ph}, the combinatorial objects representing the mentioned
evolution of data are \emph{simplicial complexes}.
These can be seen as a higher-dimensional generalization of graphs.
While graphs only have edges that link two vertices, \emph{simplicial complexes}
can link $n$ vertices and thus form edges, triangles, tetrahedrons, and so on
(for some examples, see \Cref{fig:simplicial_complex}).

\begin{figure}[h!]
    \centering
    \includegraphics[width=0.75\columnwidth]{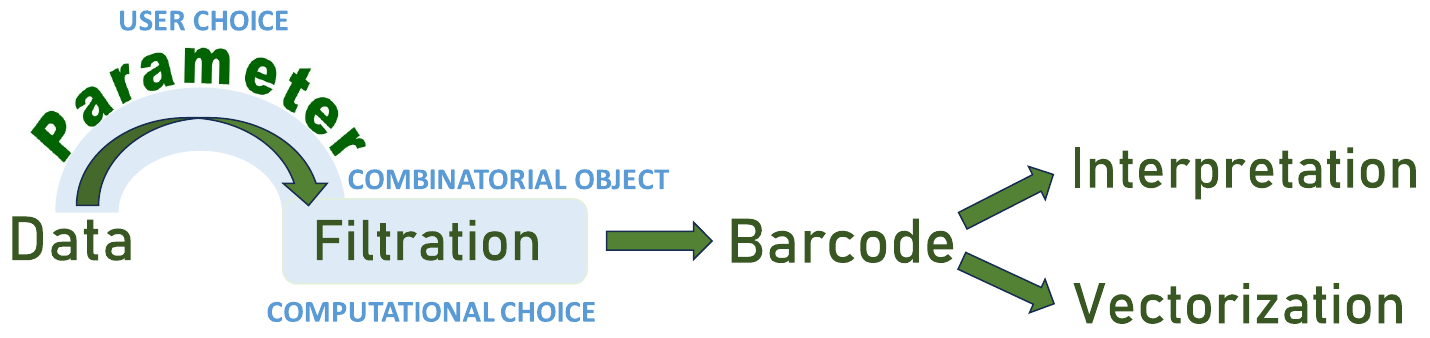}
    \caption{\gls{ph} pipeline.}
    \label{fig:ph_pipeline}
\end{figure}

Contrary to what happens for Mapper and \gls{umap}, where most of the work
goes into building the combinatorial object, for \gls{ph} this is just half
of the effort (see \Cref{fig:ph_pipeline}).
To get to the combinatorial object from the input data $X$, one first has to
choose the varying parameter $r$\footnote{For the sake of simplicity, we
restrict to the case of just one parameter; there is a rich theory for multi-parameter persistence, but the pipeline becomes more complicated and many
successful applications work with just one parameter, where we also have more
efficient algorithms.}.
Then, for each value of $r$, one obtains a space $X_r$ that results from the
input data using $r$ as the parameter value and turning it into a
\emph{simplicial complex} (we can think of it analogous to a triangulation).

\gls{ph} considers the evolution of $X_r$ with increasing parameter $r$.
Here we require that $X_r \subseteq X_{s}$, when $r\leq s$.
The sequence of $X_r$ is called a \emph{filtration}, and it is the combinatorial object investigated by \gls{ph}.

There are several methods to turn $X_r$ into a simplicial complex. 
Some, such as \textit{alpha}, \textit{Vietoris--Rips}, and \textit{\v Cech} complexes, rely on covering the data with increasingly bigger \textit{good covers}, 
i.e., sets whose intersections have nice mathematical properties, like being contractible and open or closed, to apply a \textit{nerve theorem}. 
This theoretical result, in turn, guarantees that the data and the resulting combinatorial object have the same topological properties, and thus the soundness of the analysis 
(see~\cite{unified_nerve} for a survey of formulations of the nerve theorem and its applications in \tda).
Another type of filtration, often used for building \textit{cubical complexes} in images analysis, is the \textit{sublevelset} filtration. 
Given the input $X$, one defines a function $f\colon X\to \mathbb{R}$, and then takes all the points in $X$ mapped under $f$ to a value smaller or equal the parameter $r$, for all $r\in\mathbb{R}$.
Formally, $f^{-1}((-\infty,r])=\{x\in X \ \vert \ f(x) \leq r \}$.
In the case of an image, this function is typically the average of the RGB values of a pixel. 
The choice of the filtration is sometime dictated by the input type, but in general there are several possibilities depending on the needs at hand.
We will not further elaborate on details, but this choice may affect the interpretation and usually impacts the runtime considerably.
Therefore, it is important to choose the more appropriate method for each
application to ensure efficient computation~\cite{otter2017}.

Now that we have a suitable combinatorial object, we need to extract information from it and we do so using homology.
\emph{Homology} is a fundamental concept from algebraic topology, allowing us to identify shapes that cannot be continuously deformed into each other.
An extensive treatment is beyond our scope (see for
example~\cite{edelsbrunner2010} for more details).
Informally, homology reveals the number of $k$-dimensional holes of a shape, for every integer $k$.
For $k=0,1, 2$, this corresponds to the number of connected components, tunnels, and cavities in the shape.
Crucially, given a continuous map between two shapes (or, in this case, simplicial complexes), for instance, an inclusion from $X$ into $Y$, there is a well-defined map between the holes.
Therefore, we can study the evolution of the homology in the filtration we obtained from data.
The resulting evolution of topological features can be represented as a \emph{barcode} (aka \emph{persistence diagram}), a collection of intervals (bars) that represent the lifetime of a hole within the filtration.
The length of a bar is called the \emph{persistence} of the corresponding topological feature. 
The filtration computed is stored as a matrix (called the \textit{boundary matrix}), and the barcode is 
computed via a special (i.e., with only specific operations allowed) Gaussian elimination (see, for example~\cite{bauer2017} for more details on the barcode computation).
Crucially, the barcodes generated by \gls{ph} are stable.
Therefore, we can define a meaningful distance on the data and on barcodes such that the distance between any two inputs bounds the distance between their corresponding barcodes~\cite{chazal2006}.
Additionally, also the barcode's vectorization is stable~\cite{skraba2023a}.
In practice, stability means that similar inputs result in similar outputs, ensuring that the extract information is robust against small perturbations of the input.

For a practical example, we describe the application presented in~\cite{wong2021a}.
Fine-grained segmentation of 3D images is the labeling of small components of objects.
This is crucial in object analysis such as intelligent manufacturing,
autonomous robotic manipulation, and general automatisation procedures.
To obtain correct fine-grained segmentation, it is important to identify
geometric and topological properties at different scales (e.g.\ shelves,
handles, and hinges in a closet).
Moreover, often these properties involve more complex structures than the
ones obtained by mutual distances.
The authors of~\cite{wong2021a} use \gls{ph} to enhance the prediction of a
convolution network.
The input is a 3D shape in point cloud format, and it is processed in parallel by a graph convolution network and \gls{ph},
which both output a representation in a latent space.
These vectors are then combined and augmented to obtain the augmented feature map which label the input data.

\gls{ph} produces the persistence diagrams of the loops and cavities
(1- and 2-dimensional homology, respectively).
These diagrams are vectorized using \glspl{pi}~\cite{adams2017}.
Then, to train the network, the authors~\cite{wong2021a} use a combination of
the (commonly used) cross-entropy function and a newly defined \gls{ph} loss function.
This latter ensures a small topological error, ensuring a much more precise segmentation at different scales.

As illustrated in \Cref{fig:ph_pipeline} and exemplified in~\cite{wong2021a},
barcodes can be vectorized and thus fed to a
\gls{ml} pipeline for enhanced analysis.
There is a rich theory on how to compare two datasets by comparing their
barcodes, and how to integrate \gls{ph} into \gls{ml} methods, i.e.,
kernel-based methods~\cite{kwitt2015,reininghaus2015} or neural
nets~\cite{adams2017,hofer2019}.
The well-founded theory and the interpretability of the obtained features
have contributed to the success of \gls{ph} in practice.
There are many efficient algorithms to compute filtrations and barcodes, and
to subsequently compare them, such
as~\cite{bauer2021,bauer2017,clement2014,morozov2007,tauzin2021}.

A drawback of \gls{ph} is that, unlike Mapper or
\gls{umap}, it does not easily scale to very big datasets since it boils down to Gaussian elimination on very big matrices.
On the positive side, it is more intuitive than Mapper and UMAP because usually there is a natural choice of parameter to filter.
Moreover, since it requires less parameter tuning, \gls{ph} is very apt for automation.

    \section{Survey Method}\label{sec:survey}

One of the objectives of this work is to ensure the reproducibility of the survey results.
Therefore, we decided to perform the review as an exhaustive literature
review~\cite{randolph2009}, where each step is documented and can be reproduced.
The problem defined for this survey is the review of methods from
\gls{tda} in industrial production and manufacturing.

\begin{figure*}
    \centering
    \includegraphics[width=\textwidth]{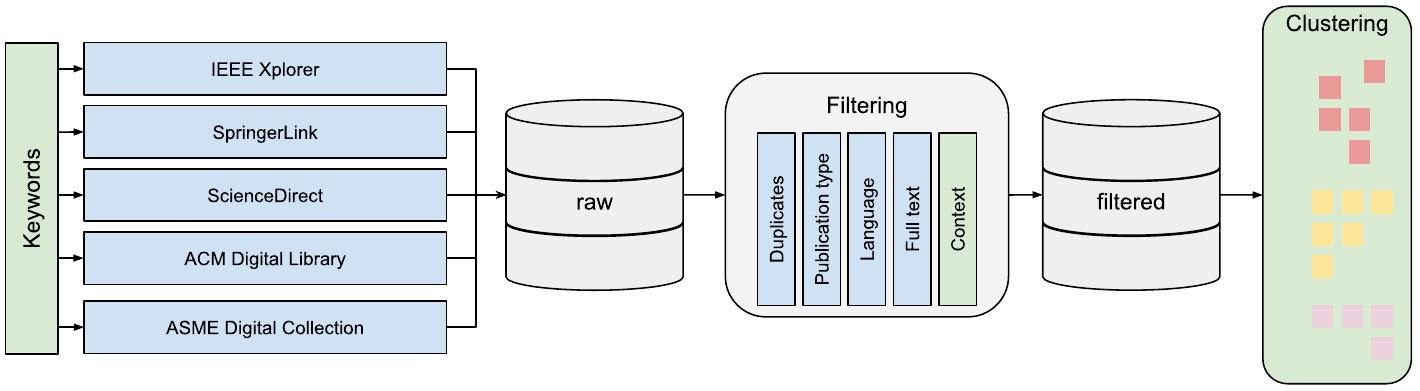}
    \caption{The data aggregation and filtering pipeline for the literature
    review consists of several steps.
    The first step is the definition of a set of keywords for the search
        (\Cref{subsec:keywords-and-queries}).
    Further, these keywords are used to search in the digital libraries
        (\Cref{subsec:digital-libraries}).
    The results are then filtered according to the criterias described
    in \Cref{subsec:filtering-results}.
    The blue boxes indicated automated steps, while the green ones indicate
    manual effort.}
    \label{fig:data_aggregation_and_filtering}
\end{figure*}

The pipeline of the survey method is illustrated in
\Cref{fig:data_aggregation_and_filtering} and contains the following steps:
definition of appropriate keywords for the search, identification of the
digital libraries where to search, and
filtering of the obtained works.
We now explain these steps in detail.

\subsection{Keywords and queries}\label{subsec:keywords-and-queries}

\begin{figure}
    \centering
    \includegraphics[width=0.6\columnwidth]{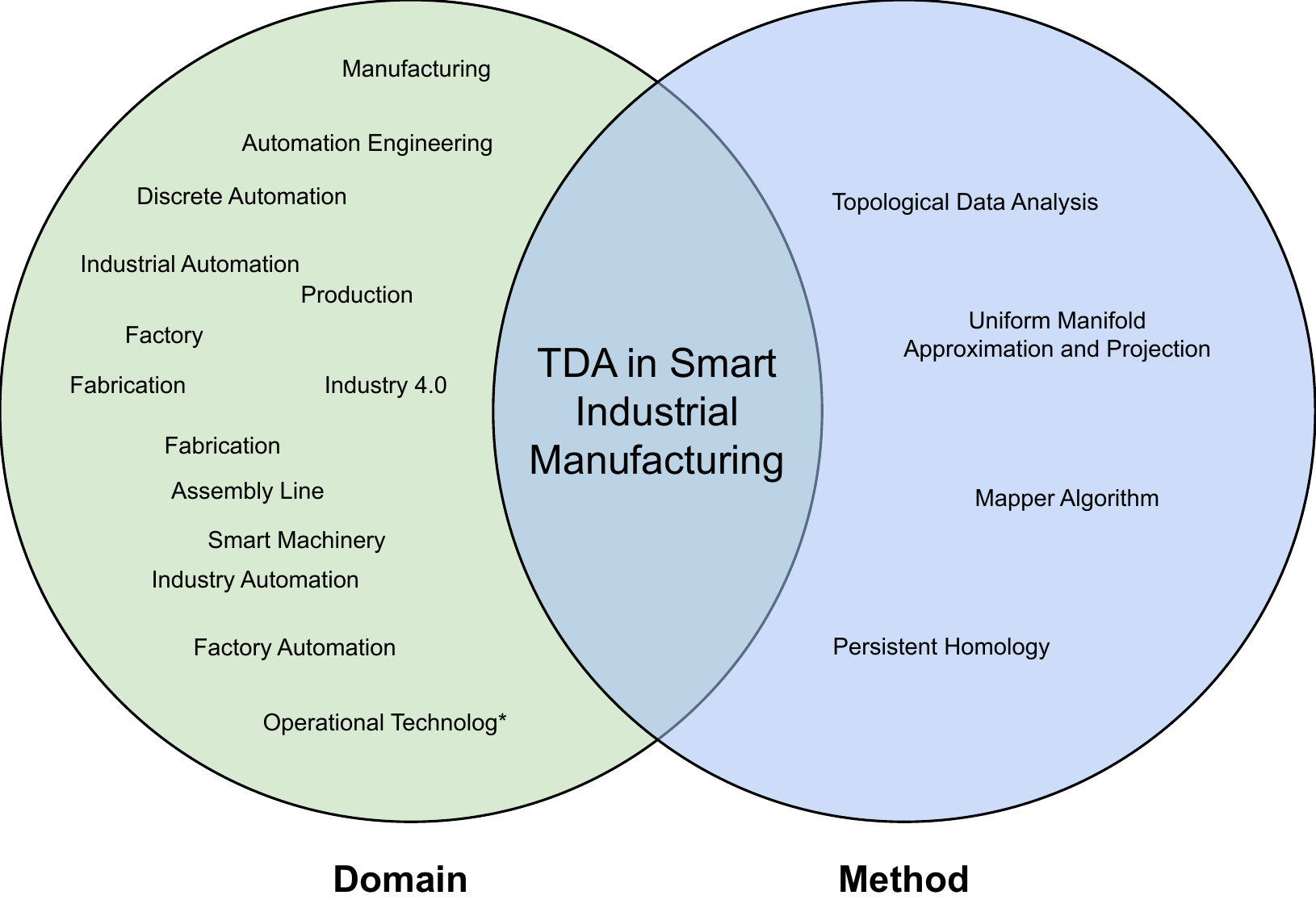}
    \caption{Venn diagram of the set of 13 keywords for the category
    \emph{Domain} (left segment) and the set of 4 keywords for the category
    \emph{Method} (right). The intersection of both sets indicates the scope
    for this literature search.
    Note: the asteristk ``*'' indicates a wildcard, so that variants of the
    keyword can be considered (e.g.\ Technologies and Technology).}
    \label{fig:venn}
\end{figure}

The search queries are defined by means of two categories, namely
\emph{Method} and \emph{Domain}.
The keywords of the categories \textit{Method} describe the \gls{tda} tools whose applications 
we are interested in finding.
On the other hand, the keywords of the \emph{Domain} category describe
applications and tasks in industrial manufacturing.
\Cref{fig:venn} illustrates these categories including the identified keywords.
The intersection of both categories is the search space for our literature
review.

The search queries are formulated depending on the search functionality of the
digital libraries.
Using an individual search query each, the results are collected and stored.

\subsection{Digital libraries}\label{subsec:digital-libraries}

We surveyed the following five digital libraries:
\begin{itemize}
    \item IEEE with the \emph{IEEE~Xplore} Digital Library
    \item Springer with \emph{SpringerLink}
    \item Elsevier with \emph{ScienceDirect}
    \item ACM with the \emph{ACM Digital Library}
    \item The American Society of Mechanical Engineers (ASME) with the
    \emph{ASME Digital Collection}
\end{itemize}

We chose these digital libraries because the first four
are the most prominent digital libraries for scientific publications in
the field of computer science and engineering, especially for the applications
of \gls{tda} and \gls{ml}.
Furthermore, these four are also the most prominent digital libraries for
scientific publications in the field of industrial engineering.
The only exception from this selection criteria here is the
\emph{ASME Digital Collection}.
It was selected because an earlier and preliminary semi-exhaustive search --
using a method similar to the one used in~\cite{tschuchnig2022} -- on a
restricted set of keywords on \emph{Google Scholar} showed additional
relevant results that are contained in the \emph{ASME Digital Collection} but not in the
first four collections.

\emph{IEEE~Xplore}, \emph{SpringerLink}, and the \emph{ACM Digital Library}
provide a search interface that allows the usage of the generated search strings.
The interfaces of \emph{ScienceDirect} and the \emph{ASME Digital Collection}
limit the length of search queries, so we split our search query into
several sub-queries and aggregate the results.
From all these digital libraries, all results are
collected and stored using the Zotero reference
manager\footnote{\url{https://www.zotero.org/}}.
For each publication its metadata is automatically extracted through its
\gls{doi}.

\subsection{Filtering results}\label{subsec:filtering-results}

For the filtering, we did not restrict the search results on a specific time
period.
The reason for this is that \gls{tda} is still a young field and the application of it in industrial
manufacturing is even younger.
The data aggregation was performed on April 23rd, 2024, and all published
works up to that date are considered.

The filtering procedure was done step-wise, removing items according to the
following criteria:
\begin{enumerate}
    \item Duplicates
    \item Relevant types of publication
    \item Language of the references
    \item Availability of a full text
    \item Context of the work
\end{enumerate}

The first step of the filtering procedure was to remove duplicates, where
duplicates are identified by the~\gls{doi} and the title of the publication.
When duplicates were found based on the title then a manual check was performed to
verify this duplicity.
Each duplicate was removed, keeping only one instance of the
publication\footnote{Here to be noted, Conference Paper with a following,
    extended Journal version are considered to be independent (since they
    have a uniqe \gls{doi}, respectively). Their relationship is indicated in the
    results section (\Cref{sec:results}).}.

In order to deliver a meaningful review, only publications with a
sufficient degree of quality were considered (i.e., peer-reviewed publications).
Based on this requirement, only publications from conference proceedings and
journals were taken into account for our survey.
While not all conference and journal publications are peer-reviewed, the majority of them are, and hardly
no status on a review process is actually provided by the digital libraries.
Other references, like preprints, presentations, books, or reports are excluded
and not further considered for our analysis.

To ensure the correct extraction of information from the publications and to be
commonly reproducible, only publications with an available English full text
were considered.
The availability is highly dependent on the access and subscriptions
of our institutions to those digital libraries.
Therefore, all publications with no available full text were screened manually, so that no
relevant publications are missed.

The semi-automatically filtered references are then analyzed for the context.
For this, all publications were screened manually.
Here, the keywords of the categories \emph{Method} and \emph{Domain} were
searched in the publications.
They had to be present in a section relevant to the contribution. 
In particular, it was not sufficient to mention any of the methods or domains only in the related work or in the outlook.

After this filtering procedure, only the remaining publications are further
considered for this survey.
In total, over $7000$ results were screened, resulting in $34$ publications that
were  considered to be relevant for our stated research questions.
These publications were then screened in detail.
Based on this screening, the publications were categorized manually into
different categories, that are presented in the following.

    \section{Results}\label{sec:results}

In total $34$ works were identified as relevant for this survey.
Each of those identified works was assigned to one of three clusters (A-C),
based on its application within the production process.
The identified clusters are:
\begin{itemize}
    \item A: Quality Control on Product Level
    \item B: Quality Control on Process Level
    \item C: Manufacturing Engineering
\end{itemize}

The resulting works are listed in~\Cref{tab:results}.
This table indicates each work with its assigned cluster (A-C) and the used
\gls{tda} method.

\begin{table}
    \centering
    \begin{adjustbox}{max width=\textwidth}
    \begin{tabular}{c c c c}
        \toprule
        Cluster & \multicolumn{3}{c}{\gls{tda} Methods}\\
        \cmidrule(l){2-4}
        & \gls{ph} & Mapper & \gls{umap} \\
        \midrule
        A &\ \cite{behandish2019,chumley2023,yesilli2021b,yesilli2022c,wong2021a,wang2022c,wang2023a,wang2024,ko2023} &\ \cite{hsu2022} &\ \cite{sarpietro2022,sessions2022}\\
        B &\ \cite{khasawneh2016,khasawneh2018a,yesilli2019b,yesilli2022a,yesilli2022b,gomez-omella2023,giri2023} &\ \cite{guo2016,guo2017} & \textendash \\
        C &\ \cite{dassisti2017,mahler2018,munoz2023,paul2023} &\ \cite{rivera-castro2019d} &\ \cite{hsu2022a,unterdechler2022,zhang2023,zhang2021c,unterberg2021,ordieres-mere2022,sansana2023,waszak2024} \\
        \bottomrule
    \end{tabular}
    \end{adjustbox}
    \caption{The results of the survey with all references, assigned to a
    cluster (A-C) over the applied \gls{tda} methods (\gls{ph}, Mapper,
    and \gls{umap}).
    A more detailed
    representation is shown in \Cref{tab:results-vs-data-matrix}.}
    \label{tab:results}
\end{table}

\begin{figure}
    \centering
    \includegraphics[width=0.8\columnwidth]{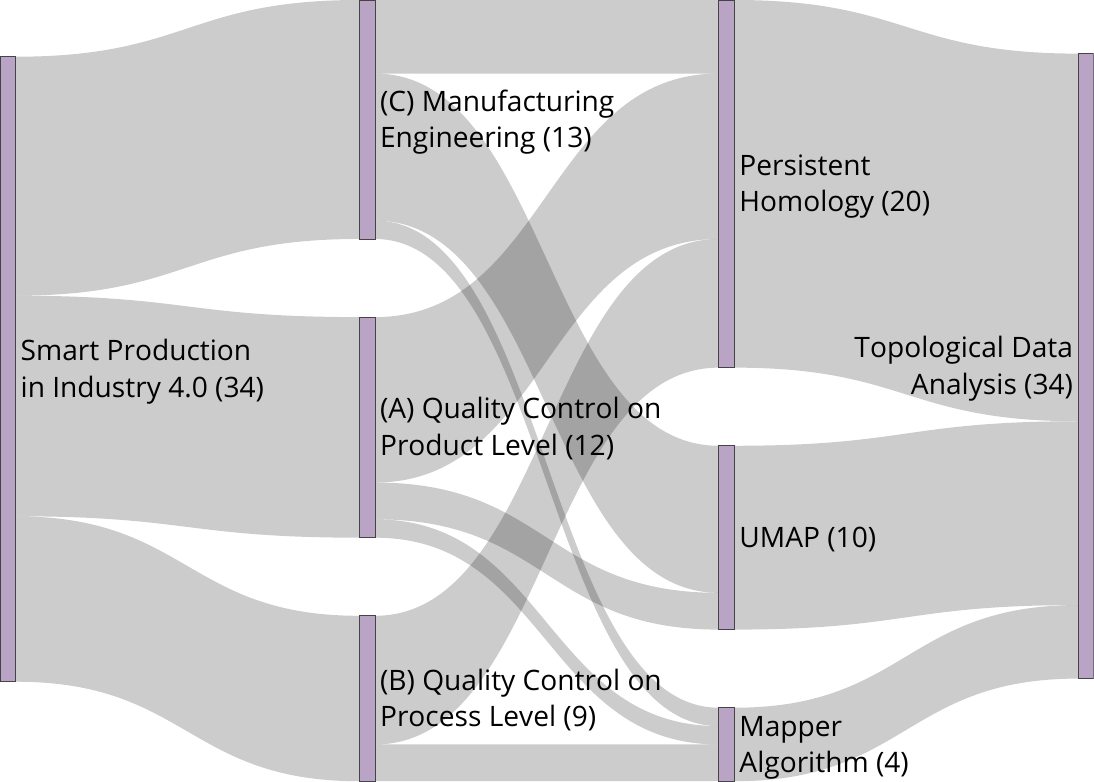}
    \caption{This illustration shows the relation of the works to the
    identified clusters. Furhtermore, the used \gls{tda} method is indicated
    for each cluster.
    The number in the parenthesis indicates the number of publications associated.}
    \label{fig:cluster-overview}
\end{figure}

\Cref{fig:cluster-overview} provides an overview of these works in relation to
their assigned clusters.
Furthermore, this illustration also indicates the used \gls{tda} method over
each application area.

\begin{figure}
    \centering
    \includegraphics[width=0.8\columnwidth]{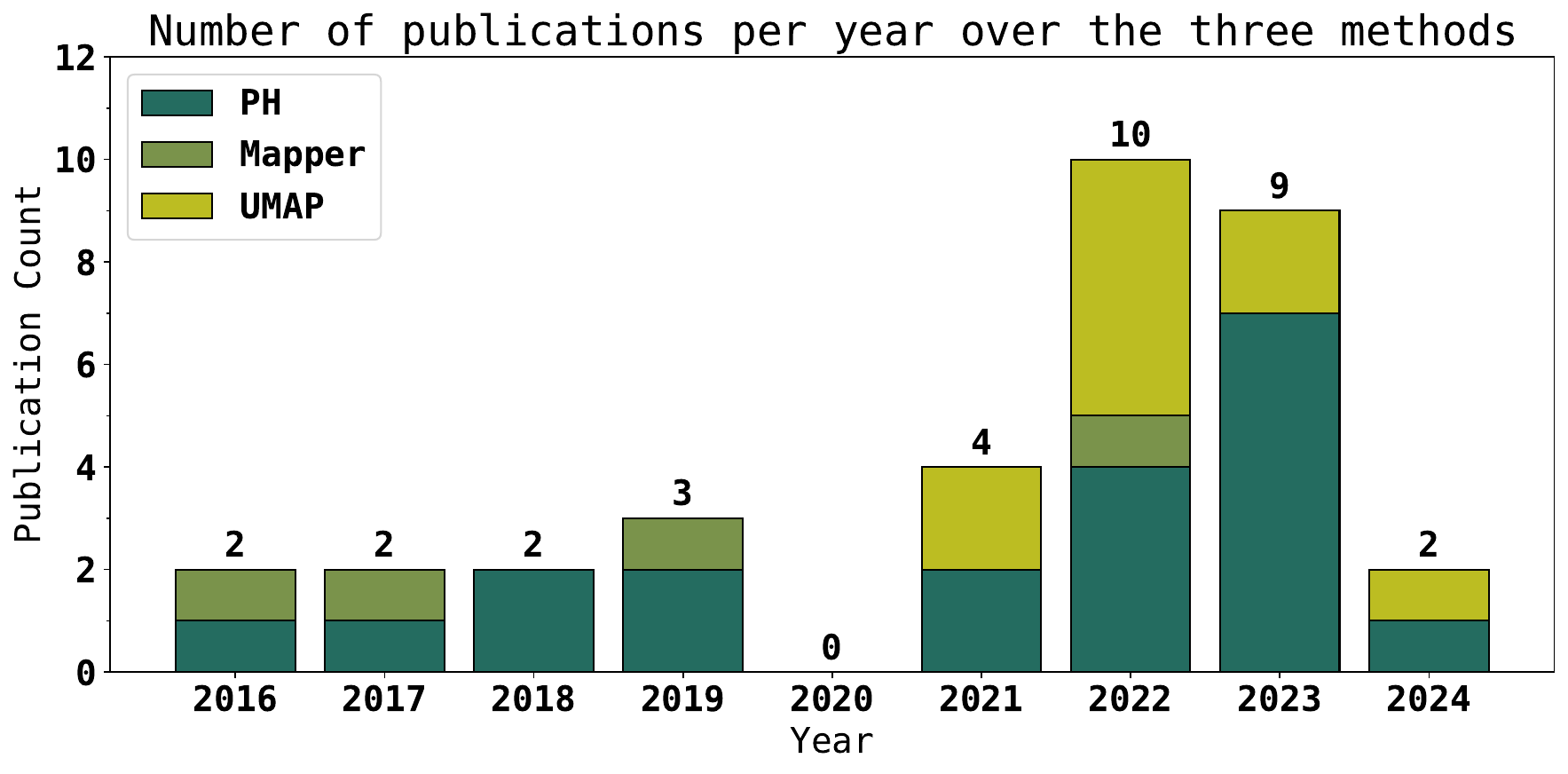}
    \caption{%
        The number of relevant publications per year.
        Counts of all methods for each year are stacked, displaying a cumulative total.
        Data aggregation was performed on April 23rd, 2024.
        Information regarding the publications is available in~\Cref{tab:results-vs-data-matrix}.
    }
    \label{fig:publication_hist}
\end{figure}

From the publication dates of the listed works can be seen that the interest
in \gls{tda} methods for manufacturing has increased over the last
years.
The first articles on the topic were published in 2016.
A major gain in interest can be observed for the years 2022 onwards.
The count of publications in 2022 is higher compared to the number of
publications in the years before 2022.
Similarly, the number of publications in 2023.
\Cref{fig:publication_hist} shows the cumulative count of relevant publications per year.
Note that the data for 2024 is incomplete, as this survey includes publications
added to the digital libraries up to April 23rd, 2024.
From this plot, we can see that the interest on \gls{tda} methods in manufacturing
has increased in the last years.
Similar trends can be observed in other fields of application, see for
example~\cite{skaf2022,corcoran2023}.

A more detailed overview of the results is given in
\Cref{tab:results-vs-data-matrix}.
This table shows each individual work, the associated cluster, the used
\gls{tda} method, and the kind of input data used to solve the task.
The type of input data is extracted from the referenced works.
The most common data type is time series data, followed by point clouds and
scalar fields.
Additionally, we found also one work employing the \gls{tda} method on textual log
files and one on a labeled graph.

\begin{table*}
    \centering
    \begin{adjustbox}{angle=90,max width=\textwidth}
        \centering
    \begin{tabular}{ccccccccccc}
        \toprule
        \multirow{3}{*}{Cluster} & \multirow{3}{*}{Work} & \multicolumn{9}{c}{Input Data Format} \\
        \cmidrule{3-11}
        & & fin.~$\subset \mathbb{R}^n$ & $\mathbb{N} \to \mathbb{R}^n$ & \multicolumn{4}{c}{Scalar Field ($\mathbb{R}^2 \to \mathbb{R}$)} & & \multicolumn{2}{c}{Graph} \\
        \cmidrule(rl){3-3} \cmidrule(rl){4-4} \cmidrule(rl){5-8} \cmidrule(rl){10-11}
        & & Point Cloud & Time Series & Wafer Map & Sign. Dist. Func. & Microscope Imag. & Surface & Logs & Funct. Block Netw. & Task Graph \\
        \midrule
        \multirow{12}{*}{A}
        &\ \cite{wong2021a}          & \gls{ph}   &            &            &          &            &          &            &&            \\
        &\ \cite{behandish2019}      & \gls{ph}   &            &            &          &            &          &            &&            \\
        &\ \cite{yesilli2021b}       &            &            &            &          &            & \gls{ph} &            &&            \\
        &\ \cite{yesilli2022c}       &            &            &            &          & \gls{ph}   &          &            &&            \\
        &\ \cite{chumley2023}        &            &            &            &          &            & \gls{ph} &            &&            \\
        &\ \cite{wang2022c}          &            & \gls{ph}   &            &          &            &          &            &&            \\
        &\ \cite{wang2023a}          &            & \gls{ph}   &            &          &            &          &            &&            \\
        &\ \cite{wang2024}           &            & \gls{ph}   &            &          &            &          &            &&            \\
        &\ \cite{ko2023}             &            &            & \gls{ph}   &          &            &          &            &&            \\
        &\ \cite{hsu2022}            &            &            & Mapper
        &          &            &          &            &&            \\
        &\ \cite{sarpietro2022}      &            &            & \gls{umap} &          &            &          &            &&            \\
        &\ \cite{sessions2022}       &            &            &            &          & \gls{umap} &          &            &&            \\

        \midrule
        \multirow{9}{*}{B}
        &\ \cite{yesilli2022a}       &            & \gls{ph}   &            &          &            &          &            &&            \\
        &\ \cite{khasawneh2016}      &            & \gls{ph}   &            &          &            &          &            &&            \\
        &\ \cite{khasawneh2018a}     &            & \gls{ph}   &            &          &            &          &            &&            \\
        &\ \cite{yesilli2019b}       &            & \gls{ph}   &            &          &            &          &            &&            \\
        &\ \cite{yesilli2022b}       &            & \gls{ph}   &            &          &            &          &            &&            \\
        &\ \cite{guo2016}            &            & Mapper
        &            &          &            &          &            &&            \\
        &\ \cite{guo2017}            &            & Mapper
        &            &          &            &          &            &&            \\
        &\ \cite{gomez-omella2023}   &            &  \ph       &            &          &            &          &            &&            \\
        &\ \cite{giri2023}           &  \ph       &            &            &          &            &          &            &&            \\

        \midrule
        \multirow{13}{*}{C}
        &\ \cite{rivera-castro2019d} &            & Mapper
        &            &          &            &          &            &&            \\
        &\ \cite{hsu2022a}           & \gls{umap} &            &            &          &            &          &            &&            \\
        &\ \cite{dassisti2017}       & \gls{ph}   &            &            &          &            &          &            &&            \\
        &\ \cite{mahler2018}         & \gls{ph}   &            &            &          &            &          &            &&            \\
        &\ \cite{munoz2023}          &            &            &            & \gls{ph} &            &          &            &&            \\
        &\ \cite{unterdechler2022}   &            &            &            &          &            &          &            & \gls{umap} &\\
        &\ \cite{zhang2023}          &            & \gls{umap} &            &          &            &          &            &&            \\
        &\ \cite{zhang2021c}         &            &            &            &          &            &          & \gls{umap} &&            \\
        &\ \cite{unterberg2021}      &            & \gls{umap} &            &          &            &          &            &&            \\
        &\ \cite{ordieres-mere2022}  &            & \gls{umap} &            &          &            &          &            &&            \\
        &\ \cite{paul2023}           &            &            &            &          &            &          &            &&  \ph       \\
        &\ \cite{sansana2023}        &            & \umap      &            &          &            &          &            &&            \\
        &\ \cite{waszak2024}         &            & \umap      &            &          &            &          &            &&            \\

        \midrule\midrule
        & \multicolumn{1}{r}{Count} & $6$         & $17$       & $3$        &    $1$   & $2$        & $2$      & $1$        & $1$ & $1$         \\
        \cmidrule(rl){5-8} \cmidrule(rl){10-11}
        & & & & \multicolumn{4}{c}{$8$} & & \multicolumn{2}{c}{$2$} \\

        \bottomrule
    \end{tabular}
    \end{adjustbox}
    \caption{All the identified works of the survey including their data format
    employed on and the method being used, grouped by the identified clusters.
    The data format is expressed in a mathematical notation and their
    occuring format within the works.
    The works on time series data are of different kind and employ the \gls{tda}
    methods to uni- and multivariate time series data.
    As scalar fields are considered Wafer Maps (also Wafer Defect Maps),
    Signed Distance Functions, Microscope Images, and Surfaces.
    Logs are considered to be heterogeneous and multidimensional textual data,
    with the necessity of preprocessing this unstructured format.
    The functional block network is a labeled graph, where the nodes are
    functional blocks and the edges are the connections between the blocks.
    The task graph is a graph, where each node represents a task and the
    edges are weighted according to the connected tasks properties.
    }
    \label{tab:results-vs-data-matrix}

\end{table*}

In the following~\Cref{subsec:cluster-1,subsec:cluster-2,subsec:cluster-3},
the identified application clusters are discussed.
For each area of application, a short description is presented, followed by a
brief summary of the associated works.
For further details on these works, we refer to the original, referenced
publications.
Additionally,\ \Cref{subsec:excluded-works} discusses works
encountered during the survey that apply \gls{tda} methods to validate analyses
performed by other methods.

    \subsection{Quality Control on Product Level}\label{subsec:cluster-1}

This cluster encompasses works focused on identifying discrepancies in
manufactured goods.
Different tasks were identified, but all share the common goal of comparing a
measured property with a reference.
During the production process, this task is performed after the production
itself, as illustrated in \cref{fig:production_process} (cf. \emph{Inspection
and quality assurance}).
In general, methods from \gls{tda} are naturally suited for the analysis of
structures, surfaces, and shapes.
We survey the usage of \gls{tda} in the context of quality control on product
level.

A natural application of \gls{tda} on the product level is the analysis of
discrepancies of the products topology, as shown in the example
in~\cite{behandish2019}.
In this work, the authors classify
topological discrepancies in additive manufacturing using \gls{tda} methods.
The products to be classified are embedded as a mesh in $\mathbb{R}^3$.
In this work, they mainly use (non-persistent) homology\footnote{For the sake of
simplicity, we account this work as an application of \gls{ph}, as
mathematically \gls{ph} can be considered to be a generalization of homology.}.
Discrepancies are detected by comparing the homological features of the
mesh with the homological features of a reference mesh.
The authors are able to demonstrate the successful application of their
method to the classification of deviations between as-design and as-produced
articles.

Another natural use case is the analysis of surfaces texture.
An early work proposes the application of \gls{ph} to
differentiate various generic surface profiles~\cite{yesilli2021b}.
Their method applies the filtration on the captured profiles, describing the
surface lattice by $0$- and $1$-dimensional persistent homology features.
In a follow-up work~\cite{yesilli2022c}, the authors apply this methodology
on the more specific task of microscope images, recorded after a {\emph Piezo Vibration
Striking Treatment}.
This treatment is a method to change the surface finish of a product, which
is very essential for e.g.\ its hardness.
With both works, the authors are able to report similar results as their
baseline methods, while eliminating the manual preprocessing steps, demanded
by traditional methods.
In~\cite{chumley2023}, the authors extended their approach to further analyze
additional properties. 
The ``tool striking depths'' and the ``roundness'' of the patterns onto the more 
general format of surface images are analyzed in addition to characterizing 
the patterns on the product's surface.

For the segmentation of shapes in point cloud data,\ \cite{wong2021a} propose to
enrich the input to an existing \gls{gnn} with topological information.
As \glspl{gnn} typically do not capture multi-scale topological information,
and misses these features in the input space.
The authors propose an approach, that enriches the ``local'' features, extracted using a \gls{gnn}
(in fact, a Graph Convolutional Network),
by ``global'' topological features.
These features are extracted using \gls{ph}, transformed to \glspl{pi} and
reshaped to a ``topological feature map''.
Both the local and the topological feature map are concatenated and fed to a
$1 \times 1$ convolutional filter to extract an ``augmented feature map''.
The augmented feature map is then used for the final segmentation task by
the \gls{gnn}.
Their \gls{ph}-based \gls{gnn}
outperforms the state of the art on methods of fine-grained
3D shape segmentation, performed on point cloud data.

A more specialized use case is the identification of defect patterns in
the wafer production.
For this task,\ \cite{hsu2022} case-study the use of Mapper.
Their method takes wafer maps as the input.
In a first step, a vision transformer extracts higher-level features from the
wafer map images.
Mapper is then applied on these extracted features.
The Mapper graph's visualization is then used to identify clusters.
the authors highlight their method's ability to identify defect patterns on
a public benchmark dataset, without a comparison to other methods.

Another application scenario is the production of electrical motors.
In~\cite{wang2022c}, the authors use \gls{ph} for the detection of eccentricity
of electric motors.
Eccentricity is a non-uniform gap between the stator bore and the rotor that
causes a non-circular rotation.
In this task, the input data is a time series of the process parameters of
the electric motors.
The time series is transformed into a point cloud in 3D, using a so-called
time-delay embedding.
On this transformed data representation, \gls{ph} is applied.
The eccentricity level can then be described by the 1-dimensional homological
features in the persistence diagram.
In their work, they are able to predict the fault levels with a reasonable
accuracy, while achieving significantly lower computational complexity.
In the follow-up work~\cite{wang2023a}, the authors compare their method to
physics-based approach, and in~\cite{wang2024} the authors provide a more
thorough theoretical description and experimental evaluation of their method.


The work in~\cite{sarpietro2022} proposes a method for the sorting of
electrical wafers.
The primary objective for their use case is to detect defects in produced wafer
maps, so that these can be discarded at an early stage.
Within the proposed Deep Learning pipeline a preprocessing step is performed.
This preprocessing applies a dimensionality reduction on the high-dimensional
input data. 
For this task, they employ \gls{umap} as a dimensionality reduction method.

A similar task is performed in a work~\cite{ko2023}, where the authors
detect defect patterns in wafers maps.
In this work, the authors propose a method using \gls{ph}.
First, a filtration is performed on the wafer maps in $\mathbb{R}^2$.
The resulting persistence diagram is then transformed into a vector
representation, using \glspl{pi}.
The neural network then performs the classification task on the vectorized
representation.
Moreover, the authors highlight the computational efficiency of the proposed technique compared to other methods.

On the task of additive manufacturing of radio-frequency devices,\ \cite{sessions2022}
propose a method for the identification of defect mechanism and their
performance impact.
This task usually requires in-line electromagnetic simulations, which are
time-consuming and expensive.
In their work, they use \glspl{cnn} on microscope images,
resulting in a vector.
\gls{umap} is then employed for the dimension reduction on the resulting
vectors.
Their approach provides to a faster and cheaper quality control process.

    \subsection{Quality Control on Process Level}\label{subsec:cluster-2}

The applications in this cluster aim to assess production process quality by
observing the state of the process.
These observables are assesed during the production process
(cf.~\Cref{fig:production_process}, \emph{Production}).
Examples of data in this context are machine states, sensor data, or other data
acquired during the production process.
From these process variables, failures, anomalies, and other quality issues can be
detected.

\begin{figure}
    \centering
    \includegraphics[width=0.6\columnwidth]{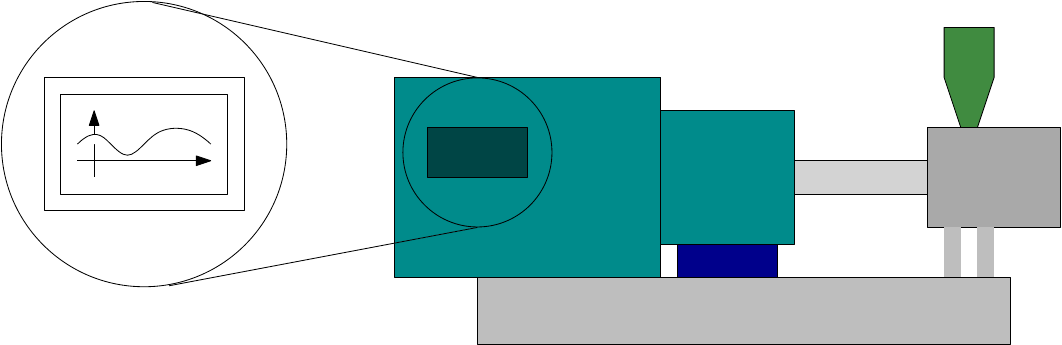}
    \caption{Applications of quality control at the process level utilize key
    process parameters, represented as time-series data, to monitor and
    control machinery processes.
    Process parameters are illustrated here on a schematic injection modeling
    machine.
    This figure is complete regarding the input data types (time-series) on
    the task of this cluster, as
    shown in \Cref{tab:results-vs-data-matrix}.}
    \label{fig:cluster-3}
\end{figure}

Nine works belonging to this cluster were found during the review.
These works deal with two major application types.
As illustrated with the example of an injection molding machine in
\Cref{fig:cluster-3}, all applications (with one exception) are based on time-series data as the input.

Given observations from key process parameters, the aim of the works
in~\cite{guo2016} and~\cite{guo2017} is to predict the productivity of a
manufacturing process.
The authors use Mapper to identify intrinsic clusters within
benchmark processing datasets.
Using Mappers output network, key process variables or features
that impact the final product quality the most are selected.
Their work showed that this model achieves the same level of prediction accuracy
as with all process variables, while being more cost-effective.

The second application within the cluster of quality control on process level is
the detection of chatter.
Chatter detection in machining has garnered attention in recent
years, as can be seen by a survey work on this particular application
domain~\cite{kounta2022}.
Detection of chatter is important since the abnormally of chattering can lead
to damage to the workpiece or the machine tools.
Using \gls{tda} methods for chatter detection has been studied in a line of
research by Khasawneh et al.

The first work~\cite{khasawneh2016} is a proof of concept, where the authors
show that \gls{ph} can be used for chatter detection in general.
Each time series is converted into a delay embedding.
The structure within this representation can then be extracted using \gls{ph}.
Abnormalities are then visible within the persistence diagram.

In the follow-up work~\cite{khasawneh2018a}, the authors propose a method for
chatter detection, based on \gls{ph} in combination with supervised learning.
Based on the previous work~\cite{khasawneh2016}, the resulting persistence
diagrams are vectorized.
The vectorized representation data is then used to train a logistic
regression classifier.
They report high accuracy in chatter detection.

In a further work
(conference~\cite{yesilli2019b} and a journal publication~\cite{yesilli2022b}),
the authors dig deeper into different supervised methods for chatter
detection based on topological feature vectors obtained using \gls{ph}.
In these works, different methods on the vectorization of the persistence
diagrams are compared and discussed.

Furthermore, in a follow-up work~\cite{yesilli2022a}, they additionally extend
the previous works, by pretraining the methods on a different source domain
(turning) and transfer the learned knowledge using transfer learning to a task
in the target domain (milling).
This work also employs dynamic time warping, to align the time series.
Their show-case's results prove that transfer learning can
be used to improve the performance of the chatter, when pretrained on a
different dataset.

The task of optimizing the production process of wafers is further addressed
in~\cite{giri2023}. 
In the previously discussed work (see \Cref{subsec:cluster-1}), the authors analyze the product itself.
In~\cite{giri2023}, they apply \tda to cluster data about produced wafers to 
analyze the root cause for defects in the produced goods by only observing latent 
data from the production process.

A number of variables are measured and recorded for each produced entity, like
sensor parameters or electrical measurements.
The corresponding high-dimensional data is then analyzed using a proprietary
toolset, which identifies homological groups during its processing
procedure\footnote{The method is described on a very high-level.
    However, similar to~\cite{behandish2019}, also this work is accounted for \ph due
    to the utilization of homological groups.}.

A challenge faced in \gls{am} is to avoid defects in the manufactured parts.
Defects can be caused by a variety of reasons, such as the porosity of the material.
The work~\cite{gomez-omella2023} compares three operational strategies
for an early stage detection and prediction of part failures due to porosity.
In this work, \ph is used to analyze process parameters (time series data).
The resulting persistence diagram is transformed into \glspl{pi},
which is then used to train a \gls{ml} classifier to
predict faulty products.
Their work concludes by noting that the usage of topological
features over basic features from time series data decreases performance in
all experimental evaluations.
However, this work does not report any experimental evaluation on the
combination of both, topological and basic features.
Concluding, the authors discuss that the data is incomplete, as these lack
some parameters, and the experiments are hard to replicate, due to varying
ambient conditions.
These results therefore need to be interpreted with caution.

    \subsection{Manufacturing Engineering}\label{subsec:cluster-3}

This section covers the applications of \gls{tda} in the field of
manufacturing engineering.
With manufacturing engineering, we refer to the engineering discipline that
designs, analyzes, and improves manufacturing processes and systems.
The tasks in this area are not centered on the products per se, but rather they are
concerned with the overall processes and systems that are used to manufacture
these products.
These tasks are included within the full manufacturing engineering process
(cf.~\Cref{fig:production_process}) and beyond.
Tasks of manufacturing engineering include the optimization of the material
flow, the optimization of the production process, and the optimization of the
production system, as well as the selection of components and the design of
production lines.
For this kind of application, $13$ works were found in the literature.

A very common task for a manufacturer is the temporal planning of the
production.
The demand for a product can vary depending on several factors.
Such factors can be the season, the location, the weather, or other events,
like promotions or holidays.
Not meeting the demand can lead to a loss of customers, while overproduction
leads to a monetary loss, e.g., caused by the massive storage or disposal of the products.
Depending on the use case, it may be beneficial that similar products are
grouped and share forecasting models.
In~\cite{rivera-castro2019d}, the authors propose a method for demand
forecasting with different techniques.
In order to generate forecasts for a new product, a predictor-model needs to
be selected.
For this selection process a $k$-nearest neighbor algorithm based on a Mapper Graph is proposed.
By using the topological properties of the historic time-series data, the authors conclude
that the selection of the predictor-model is more accurate and much faster than other methods.

The recent work~\cite{munoz2023} addresses the problem of relying on expert knowledge and experience of machine operators.
Turnovers on machinery require re-parameterization of the machines.
However, these changes of parameters are based on the experience of the machine operator.
This poses the drawback that this re-parameterization is only reproducible up to
a certain degree and an operator needs to be trained for a long time in order to
gain the required experience on the particular machine type.
In their work, the authors propose a strategic synergistic use of existing \gls{ml} tools
to extract a reduced manifold from existing geometric designs of signed
distance functions.
Using interpolation techniques, the reduced manifolds are used to generate
new geometric designs by inferring missing information using clustering techniques.
Their work heavily relies on \gls{ph} and \gls{pi}.

Material flow optimization is the task of optimizing the schedule and the flow of
material through a production system.
This task involves transporting raw material from the depot to the
production, transporting semi-finished products between the production
lines, and transferring finished products from the production to the
depot back again.
Given the complexity of the production systems, the optimization of the
material flow is a challenging task, since all involved components have
different capacities, changeover times, and other constraints.
Having these constraints in mind, the optimization of the material flow
must be covered from a business perspective, as well as from a technical
perspective.
The task of material flow optimization is a multi-objective optimization
problem, where the objectives are to minimize the costs and the time of the
material flow.
A visual benchmark is proposed in~\cite{dassisti2017}.
For this benchmark task, the material flow from depot to the production line is to
be optimized as a multi-vehicle routing problem, with the data embodied as a
point cloud in a multidimensional space.
The evaluation is performed using \gls{ph}.

An optimization task of another kind is presented in~\cite{mahler2018}.
Given independent moving objects, like grippers or robots within a production
environment, these must be secured against collisions with other objects, the
surrounding environment, and, most important, human operators.
Securing these object can be achieved through physical cages,
where the objects operate within.
However, these cages can be very complex to built, are inflexible and expensive.
A more cost-effective solution is to use virtual cages, where the objects are
restricted by a virtual boundary.
The task of this work is the syntheses of planar energy-bounded cages that have an optimal configuration for a given object.
By means of identifying gripper and force-direction configurations and the
application of \gls{ph}, an optimal configuration is found.
For this purpose, the objects and grippers are modeled as point clouds.

Making sure that a product leaving the production line is of a certain quality
is a major task in manufacturing engineering.
Releasing faulty products can lead to a loss of reputation and, in the worst
case, to a loss of human life.
To make sure to meet the quality requirements, system-level tests are performed on
each product.
For the generation of rules to classify the failed parts,\ \cite{hsu2022a}
proposed a method that employs \gls{umap} for dimensionality reduction.
Unlike the applications discussed in \Cref{subsec:cluster-1}, this work does
not focus on the detection of discrepancies; rather, it concentrates on
generating rules that allow the classification of manufactured parts.
The write-up of the paper leaves out details of their proposed solution, but
the main ideas seems to be that they embed the information about passed and failed tests
in a multidimensional space and then use \gls{umap} to reduce the
dimensionality.
This representation is then used to generate rules for the classification of
failed parts.

\gls{ot} systems are Cyber-Physical Systems, i.e., where the physical
part is controlled by a computer system through sensors and actors.
Establishing and extending such systems is a challenging task, as these systems
tend to become highly complex and heterogeneous.
Identifying repeating patterns allows for the optimization of resource
allocation and system configuration, enhancing operational efficiency and
reliability.
To find repeating patterns in these systems,\ \cite{unterdechler2022} propose
a method to reuse components that are already established in the system,
based on labeled graphs.
This guarantees higher reliability, lower cost, and lower effort for
maintenance.
For their method, they use \gls{umap} for dimension reduction.

In \indfour, the security of production systems, and \gls{ot} systems in
general, is a major concern.
\gls{ot} has far different requirements than \gls{it} systems.
\gls{ot} systems, unlike \gls{it} systems, directly interact with the physical
world, necessitating robust security measures that account for both digital and
physical vulnerabilities.
This also reflects the challenges in the field of security~\cite{ervural2018,stouffer2022}.
Covering the topics on \gls{ot} security there is only one work found during the survey.
In~\cite{ordieres-mere2022}, the authors propose using \gls{umap} for downstream tasks in security
applications with steel production processes, whereas the data is given as
time-series data.

A case study in the herbal medicine manufacturing industry is presented
in~\cite{zhang2023}.
In this work, the authors attempt to analyze the degradation of the evaporation
process, since this process is a major cost factor in the production.
For the analysis, they employ \gls{umap} for dimension reduction on time-series data.

For the task of preventive maintenance,\ \cite{zhang2021c} propose a method for
the analysis of machine maintenance data.
Such datasets are often heterogeneous and multidimensional logs.
These data need to be analyzed to find patterns that indicate a failure.
In their work, the authors introduce a visual analytics approach for the
diagnosis of such heterogeneous and multidimensional machine maintenance data
(textual log data),
where \gls{umap} is used for dimension reduction as one step in the processing
pipeline.

Another approach for the analysis of machine maintenance data is presented
in~\cite{unterberg2021}.
In their work, the authors mitigate the problem of the degradation of the machinery in
the fine blanking industry by observing the wear of the machinery using
acoustic emissions.
This approach is based on \gls{umap} and hierarchical clustering on time-series data.
The authors claim that the data visualized in two dimensions resemble the
temporal dependence of the data, while allowing to identify the wear of the
machinery.

A use-case on the planning of a multi-robot collective transport tasks
is described in~\cite{paul2023}.
This work facilitates \textit{topological abstractional} features for graph
reinforcement learning.
In their work, they replace the graph laplacian (as needed by their base method)
with a laplacian matrix that is computed using \ph.
This laplacian matrix incorporates higher-order topological information from the
input task graph, which is then used as an input for the policy network.
The evaluation reports similar performance to state-of-the-art methods,
while the necessary computational time was reduced significantly.

The task of monitoring and analyzing the degree of degradation of machine
equipment is a crucial step in smart manufacturing.
An optimal strategy may prevent machine failures and the overall possible
operational time may be maximized.
In~\cite{sansana2023}, the authors present a method for the analysis of machine
degradation in batch production.
In their work, they employ \gls{umap} to cluster time-series data as
one step in their analysis pipeline.

The work~\cite{waszak2024} discusses the tasks of detecting critical events
in machinery of mining and metal processing.
Critical events may be characterized as patterns within the data, that indicate
a potential failure of the machinery.
This work proposes the equipment of vibration sensors on the production machinery to
record vibrations of the machinery.
The data may then be used to improve maintenance strategies, minimize
downtimes, or improve production throughput.
In their use-case, the acquired data is fourier-transformed and enriched by
information from \textit{Machine Execution Systems} and \textit{Enterprise Resource Planning} Systems.
The resulting set of features is clustered using \umap.

    \subsection{Excluded works}\label{subsec:excluded-works}
Besides being used for data analysis tasks on input data, \gls{tda} methods are
also used for validating the analysis performed by other methods.
For example, in~\cite{fujita2017} and~\cite{howland2023a}, \gls{ph} is used for
evaluation, and, in~\cite{cooper2023,lu2021,tercan2022,wang2023,zhou2023},
and~\cite{zhu2021}, \gls{umap} is used for the same purpose.
Since this survey addresses the direct applications of \gls{tda}, these works
are not included in the results and the remaining discussion.

Furthermore, other articles mention the use of \gls{tda} methods on manufacturing
systems in their work, but did not do any empirical work.
Examples for this are the potential application of \gls{tda} methods to hybrid
twins for smart manufacturing~\cite{champaney2022} and 3D
printing~\cite{wang2020d}.
    \section{Discussion}\label{sec:discussion}

\subsection{Interpretation of the Results}\label{subsec:interpretation}

As we showed, the number of industrial applications of \gls{tda} is rising
over the last years, but the number of applications is still low (c.f.~\Cref{sec:results}) compared
to more established domains, like biology (81) or medicine (84)
(numbers according to DONUT~\cite{giunti2022}).
This may result from several factors.
First, the discrepancy of focuses: practitioners in industry are primarily
engineering-oriented, while \gls{tda} experts \gls{tda}
are more research-focused.
Second, the integration of other data-driven methods whose use requires less background, such as \gls{ml},
into production systems is still in its infancy, and therefore has not yet reached its limits.
Third, the lack of mutual awareness.
Practitioners in industry are not yet fully informed of the potential of \gls{tda}, and 
\gls{tda} experts are not aware of attractiveness of the challenges the industrial and manufacturing
setting pose.

Method-wise, we observe a disparity in utilization.
The most predominantly utilized method is \gls{ph} (20 publications).
This prevalence might be caused by its interpretability compared to the other methods.
The dimension-reduction method \gls{umap} is used in 10 referenced works.
All of them underscore \glspl{umap} topology-preserving
characteristics, favoring its use over other dimension-reduction methods.
The method used the least is Mapper with only 4 works.
We speculate that a reason for this is that Mapper has been patented and
commercialized~\cite{carlsson2010}, possibly preventing the academic community to freely study its applications.
At the same time, the need to protect a \tda method proofs its
commercial value.

\umap and \ph are the methods that are applied on the most diverse data types.
\ph is utilized on time-series data, point clouds, and scalar fields.
Likewise, also \umap is used on a variety of data types, including time-series
data, point clouds, and even log data.
Interestingly, Mapper is currently applied to time-series data
only, except one work dealing with scalar fields.

Unlike the others, \Cref{subsec:cluster-2} mostly (with one exception) leverages time series data.
This is rather naturally caused by the manual clustering since the process level is the level where time series
data is most prevalent.
However, the information about the process itself can be also embedded in a
different format, e.g., as summary information of a production process,
similar to what is being offered by the OPC~UA
specification~\cite{euromap77} for Injection Moulding Machines to MES systems.
In this case, the process information could also be embedded as a graph or a
point cloud.
This approach can be observed in the only, relatively recent work that is
not dealing with time series data.

In each of the three identified application cluster, the \tda methods utilize
different characteristics of the input data.
The works on Quality Control on Product Level (\Cref{subsec:cluster-1}) are
mainly on point cloud data.
On this kind of data, \ph is used to extract topological features and
outperforms traditional methods, due to its ability to capture the data's
topological structure over multiple scales, while also being robust to noise (see \Cref{subsec:ph}).
For cluster B, described in \Cref{subsec:cluster-2}, mostly two characteristics of the data are leveraged:
time series data are converted into a delayed embedding that reveals a hidden,
yet different topological perspective , and Mapper is used to cluster the data according
to its topological structure (see \Cref{subsec:ph,subsec:mapper}, respectively).
Further, the application of \ph in the field on chatter detection
reduces the influence of irrelevant noise.
Within the cluster of Manufacturing Engineering (\Cref{subsec:cluster-3}), the most used method is topological
aware dimension reduction.
Further, also in this case, Mapper reveals latent structure of high-dimensional
process data (see \Cref{subsec:mapper}).
Additionally, \ph is leveraged for its ability to automate processes by
considering the full value range of a parameter and being robust to noise (see \Cref{subsec:ph}).

The application of \umap is absent in Cluster B, where no studies
implementing dimension reduction have been identified -- Mapper is instead used
for clustering purposes.
However, exploring dimension reduction techniques, such as \umap, could prove
beneficial in the context of multivariate time-series data, too.

\subsection{Opportunities and Challenges}\label{subsec:pros-cons}

We now list some pros and cons of the described \gls{tda} tools.

\subsubsection*{Opportunities for Application}
\umap helps against the curse of dimensionality, a prevalent
challenge in data analysis characterized by the sparsity of data in
high-dimensional spaces.
Such kind of data can very likely be observed on production processes, where
many sensory inputs are recorded (e.g.\ \cite{euromap77}).

\ph provides novel techniques to automate analytical processes by considering the full value range of a parameter.
This approach is highlighted, for example, in the findings of a survey presented
in~\cite{casolo2022}. 
An example from the conducted survey is the detection of changes is a time series
signal that highly depends on the value of the underlying signal, as
with~\cite{kounta2022}.

\umap and Mapper can deal with large datasets, a particularly
valuable property in industrial settings, where data are usually big
(e.g.\ \cite{euromap77}).

In general, \tda methods are more transparent than, for example, \ml, where the
analysis happens in a ``black box''.
Businesses typically prefer methods that are reliable over those that might
offer better accuracy but are less understandable.
From a strategic standpoint, this makes \tda a compelling choice for the application in
industrial settings, where explainability is often necessary for, e.g., quality
control or certification purposes.

Lastly, \tda software are usually open source and thus very accessible.
With versions available in programming languages such as Python, R, and C++,
\tda methods can be applied almost off-the-shelf, without a deep understanding of the theoretical
underpinnings. 

\subsection*{Opportunities by the Complementarity to other methods}
Compared to other well established methods, \tda can overcome limitations in
areas that are considered to be ``solved''.
One prime example are \glspl{cnn}, where, depending on the fixed size of the
kernel, only local information is extracted.
Global information, like the shape of the captured object, is not directly
accessible.
Another noteworthy example are \glspl{gnn}:
Graphs and simplicial complexes share characteristics, as both can be considered
as special cases of hyper-graphs~\cite{bodnar2021a,zhang2023a}.
While \glspl{gnn} have already shown significant success in various
applications~\cite{kipf2017,zhang2018},
they are - by their design - inherently limited to pairwise interactions between
nodes and fail to recognize higher-dimensional features, such as loops or
cavities within graphs.
With both examples, when leveraging a filtration procedure, as used in \ph, this
limitation can be overcome:
by varying a parameter under observation over multiple scales, the topological
information is extracted continuously from a local to global observation point
of view.
For a more detailed discussion on the relation of \glspl{gnn} and \tda, we
refer to~\cite{hensel2021,hofer19a}.

Moreover, \tda, by its very nature, extracts or uses information about shapes and topological features
that no other data analysis method can find.
Such an application is exemplified in~\cite{behandish2019}, where changes in
topological structures are detected in additive manufacturing.
%
However, one does not need to restrict to \tda, and it can be seamlessly
integrated with other methodologies.
An illustrative example is its combination with techniques from \ml
(like Support Vector Machines, \glspl{cnn}, or Transformers),
where
topological features are extracted using techniques from \tda.
According to~\cite{hensel2021}, the application of \tda with \ml can be
grouped into two main categories, namely intrinsic- and extrinsic topological
features for \ml.
Intrinsic topological features incorporate topological analysis of the
machine learning model itself and influence the model's architecture or training.
Interestingly, none of the identified works of this survey is of this category.
We see this gap as potential opportunity for future \tda research in this
application domain.
Extrinsic topological features, however, enable the use of topological
features extracted from a given dataset.
This is enabled by vectorized persistence diagrams, like \gls{pi}~\cite{adams2017} or specialized kernels~\cite{reininghaus2015}.
The degree of facilitating the topological information can vary and highly depends on
the use-case and data:
\begin{itemize}
    \item Topological features are sufficient on its own, i.e., naive and direct
    analysis of vectorized representations of persistence diagrams.
    Here, no other features are used.
    An example from this survey is the supervised analysis of chatter on
    time-series signals~\cite{yesilli2019b}.

    \item Adding topological information to existing pipelines.
    Information about shape is often informative and topological features can be used as
    complementary features to improve existing methods.
    This can be accomplished by, e.g., concatenating topological features to the
    existing, classical input features.
    The work in~\cite{paul2023} illustrates this on a use-case by adding
    topological features to an already existing method and reporting an improved
    performance.
\end{itemize}
Another noteworthy approach to combine \ml and \tda is that \ml methods
preprocess data, so that topological aspects can be derived from the
resulting representation.
As an example,\ \cite{hsu2022} facilitate a vision transformer to extract
higher-level
features from wafer maps, which are then analyzed using Mapper.

\subsubsection*{Open Challenges in \tda}\label{subsubsec:challenges}
Nonetheless, there are many open challenges in \gls{tda}.
For example, despite its success, and unlike Mapper and \umap, \ph does not scale very well
to large datasets.
This is particularly challenging since the method itself is hard to parallelize.

Moreover, many applications have several parameters that need
to be considered simultaneously, and the so-called ``multiparameter persistence''
is starting only now to have viable software~\cite{botnan2023}.

In addition to these limitations, there are also some
challenges in the application of \tda in the industrial setting.

\subsubsection*{Challenges in Application}
Although implementations of Mapper are freely available, its application faces a
significant obstacle due to patent protection~\cite{carlsson2010}.
This legal constraint could hinder its adoption in industrial and commercial
settings, unlike other methodologies that do not encounter legal barriers.

Furthermore, while methods such as \gls{ml} can be applied in a problem-agnostic
manner, the use of \tda methods require a more discerning approach.
A deep understanding of the data and the specific problem at hand is crucial for
employing topological methods effectively, making them less straightforward.

Another difficulty is the selection of the appropriate \tda tool for the task at hand.
The descriptions and discussion in \Cref{sec:tda} and \Cref{sec:results} help mitigate this issue.
To further address it, we provide some guidelines tailored to industry experts to select the appropriate \tda method based
on their specific data type or analysis requirements.
These guidelines also offer an overview of some possible
future research directions. 
We are not aware of a similar list of guideline.

\begin{enumerate}
    \item \textbf{Time Series Data:} For analyzes involving time series, such as industrial process data (see \Cref{subsec:cluster-2}), \ph is
    recommended due to its effectiveness in capturing temporal topological
    features~\cite{perea2019}.
    For dimension-reduction of multivariate time-series data, \umap is
    advisable, as it preserves the topological structure of the data~\cite{ali2019}.

    \item \textbf{Point Cloud Data:}
    In higher ambient dimensions, the practical feasibility of computing \ph with current implementations is typically limited to datasets containing $10^2$ to $10^3$ points (see the discussion on open challenges in \tda, \Cref{subsubsec:challenges}).
    This limitation is particularly significant when analyzing higher-dimensional topological features, such as cavities, rather than just low-dimensional features like connected components and loops~\cite{bauer2021, bauer2017, otter2017}.

    If the point cloud is in low ambient dimension (up to 3; e.g., 2D images or 3D scans, like reported in \Cref{subsec:cluster-1}),
    then \ph, with the so-called \emph{alpha complex}, is efficient also for bigger datasets ($10^4$ points)~\cite{carlsson2023}.
    For larger datasets, Mapper or \umap should be considered for dimension reduction first.
    The choice between them should be guided by the specific characteristics and requirements of the data.

    \item \textbf{Automation of Processes:} For the automation of tasks or processes,
    anytime there is a clear choice for a 
    varying parameter (e.g.\ analysis of signed distance functions or task graphs in \Cref{subsec:cluster-3}) \ph should be considered~\cite{edelsbrunner2013}.

    \item \textbf{High-Dimensional Data:} For data residing in high ambient
    dimensions, \umap is the preferred method to mitigate the curse of
    dimensionality~\cite{mcinnes2018}.
    Sample applications can be found in
    \Cref{subsec:cluster-1,subsec:cluster-3}.
\end{enumerate}

This list is an introductory, practical guide for industry professionals. 
It is incomplete and possibly suboptimal once one is more familiar with \tda tools, 
but it gives the beginner very useful rule-of-thumb to approach the field.

We hope, moreover, that it will encourage practitioners in areas where \ph,\ \umap,\ or Mapper have yet
to be used to test these methods.
We anticipate that several tasks can be analyzed with \tda methods, resulting novel insights in these areas. 
Even problems considered ``solved''
(refer to~\cite{krauss2023} for examples and datasets) can be re-examined and
further explored.

\subsection{Future Research Directions}\label{subsec:further-research}

The realm of \indfour describes the utilization of advanced industrial analytics
for adaptive decision-making.
Despite this focus, this survey has not identified any studies on complex
decision-making tasks such as automated control or data-driven decision-making.
However, achieving the objective of adaptive manufacturing systems requires
sophisticated methods.
We claim that \tda can be effectively utilized for this purpose, thereby
facilitating the realization of smart manufacturing systems.

In the setting of industrial production and manufacturing, the scale and physical
nature of data are crucial aspects.
This contrasts with fields like finance, where data is predominantly abstract,
encompassing diverse physical measurements such as pressures, temperatures, and
positions.
Consequently, transforming features in this industry can significantly affect
the data's topological and geometrical characteristics.
Nevertheless, employing methods that preserve topology, such as \umap, to
reduce data dimensionality can maintain the integrity of the data's topological
and geometrical structure.
This strategy is particularly promising for handling multi-variate time series
data common in industrial production processes.

\tda can also be applied on data that has been transformed by
the Fourier Transform.
Analyzing data in its Fourier-transformed representation is a common practice
across various industrial sectors outside \tda, including control theory and
image processing.
To the author's knowledge, there is limited literature on this specific
application within the \tda field.
The only identified publication is by Huber, which presents a toy example
focusing on the feedback optimization of a closed-loop
controller~\cite{huber2021}.

As discussed in \Cref{subsec:interpretation}, the application of \tda in
industrial production and manufacturing is still in its early stages.
Thus, there is a substantial opportunity for cross-disciplinary
collaboration between the fields of production, manufacturing, and \tda.
Engaging in such collaborations is expected to generate novel applications and
methodologies, providing benefits to both areas.

Lastly, we would like to see, with a few years of distance, a recurring survey
on the same topic as this work.
This would allow to observe the development of the application of \tda in
this domain and give insights into the development of the field.
Given the nature of the method used in this work, this is highly reproducible
and open for other researchers to extend this work.

\subsection{Limitations of this Study}\label{subsec:limitations}
The results of this study are biased towards the chosen method.
First, the application of \gls{tda} in smart manufacturing is a growing field.
Thus, this survey can only be seen as a snapshot of the current published
scientific literature.
The search terms for this study are intentionally broad, followed by a
very restrictive manual filtering procedure.
However, these search terms may not capture the full literature available.
Moreover, since we restricted to some publishers and databases, some publications may have been missed, although a semi-exhaustive preliminary search was conducted to
mitigate this bias.
Furthermore, gray literature, like pre-prints, is also ignored for this study.

We also want to highlight that this study only considers works in industrial
manufacturing.
There are other industrial fields that may also interest this
article's target audience.
There are, for example, works on process optimization in oil, gas, and chemical
industry that employ methods from \gls{tda} (see~\cite{casolo2022,liu2023,nilsson2022}).

Lastly, we want to emphasize that not all industrial applications are
published in academic literature.
Revealing the techniques employed on proprietary issues frequently results in
a competitive edge. 
In the authors' experience, these applications are sometimes published only after a significant delay, if at
all.

    \section{Conclusion}\label{sec:conclusion}

This study contributes to the existing body of knowledge in several aspects:
\begin{enumerate}
    \item We provide an overview of the state of the art in the
    application of \tda within industrial and manufacturing systems;

    \item Based on the literature, we outline the areas where \tda methods have
    been implemented successfully, showcasing the diversity and potential
    of \tda in addressing complex industrial challenges;

    \item The identified application scenarios are thoroughly analyzed,
    revealing existing gaps and potential future research directions, not
    confined to the presented scenarios but applicable in a broader context;

    \item Finally, we provide a set of guidelines to aid practitioners and
    researchers in applying \tda within the field of industrial production and
    manufacturing.
\end{enumerate}

For the survey, a transparent and rigorous methodology was employed to search
for and identify relevant literature, ensuring the reproducibility of our findings.
This approach yielded $34$ works, which were analyzed.
These studies were categorized into three distinct groups:
\emph{Quality Control at the Product Level},
\emph{Quality Control at the Process Level}, and
\emph{Manufacturing Engineering}.
Each study is reviewed, focusing on the data formats used and the specific
\gls{tda} methods applied to each case.

Through this research, we demonstrate that \gls{tda} is exceptionally
well-suited for analyzing complex datasets derived from sensors and other devices
within the realm of industrial production and manufacturing.
Additionally, we highlight that the application of \gls{tda} in this domain
is still nascent, presenting significant potential for future research.
To address the challenges of integrating \tda into industrial production, we
propose a set of guidelines to aid practitioners in applying
\tda within this field.

The evolution of \gls{tda} within the context of industrial production and
manufacturing is at a preliminary stage.
Despite existing challenges, we are optimistic about the substantial
potential of \gls{tda} in this domain.
This work serves as an initial step towards raising awareness of this
potential.
Looking ahead, we envision a robust collaboration between academia and
industry to nurture the field's growth and expand the range of applications.

    \section*{Acknowledgements}
    Martin Uray and Stefan Huber are supported by the Christian Doppler Research
    Association (JRC ISIA).
    Barbara Giunti and Michael Kerber were partially or fully supported by the
    Austrian Science Fund (FWF) [P 33765-N].



    \section*{Author contributions}
    \uram{Did I forget something? Did I do this correct?}
    \textbf{Martin Uray}: Conceptualization, Methodology, Software,
        Validation, Data Curation, Writing - Original Draft,
        Writing - Review \& Editing, Visualization, Project administration.
    \textbf{Barbara Giunti}: Writing - Original Draft, Writing - Review \&
        Editing, Visualization.
    \textbf{Michael Kerber}: Writing - Original Draft,
        Writing - Review \& Editing.
    \textbf{Stefan Huber}: Writing - Original Draft, Writing - Review \&
        Editing, Funding acquisition, Supervision.

    \bibliographystyle{elsarticle-num}
    \bibliography{survey_journal_tda_industry,results}
\end{document}